\documentclass[sigconf]{acmart}

\usepackage{hyperref}
\usepackage{multirow}
\usepackage{colortbl}
\usepackage{amsmath}
\usepackage{pifont}
\usepackage{balance}

\definecolor{darkgreen}{rgb}{0.0, 0.2, 0.13}

\AtBeginDocument{%
  }

\copyrightyear{2023} 
\acmYear{2023} 
\setcopyright{acmlicensed}\acmConference[MM '23]{Proceedings of the 31st
ACM International Conference on Multimedia}{October 29-November 3,
2023}{Ottawa, ON, Canada}
\acmBooktitle{Proceedings of the 31st ACM International Conference on
Multimedia (MM '23), October 29-November 3, 2023, Ottawa, ON, Canada}
\acmPrice{15.00}
\acmDOI{10.1145/3581783.3611767}
\acmISBN{979-8-4007-0108-5/23/10}

\begin{document}

\title{Beyond First Impressions: Integrating Joint Multi-modal Cues for Comprehensive 3D Representation}

\author{Haowei Wang}\authornote{Work is done during internship at Fuxi AI Lab, Netease Inc.}\authornote{Equal Contributions.}
\email{wanghaowei@stu.xmu.edu.cn}
\affiliation{
  \institution{Key Laboratory of Multimedia Trusted Perception and Efficient Computing,
  Ministry of Education of China,
  Xiamen University}
  \city{}
  \state{}
  \country{}
}

\author{Jiji Tang}\authornotemark[2]
\email{tangjiji_bupt@163.com}
\affiliation{%
  \institution{Fuxi AI Lab, Netease Inc.}
  \state{}
  \country{}
}

\author{Jiayi Ji}\authornotemark[2]
\email{jjyxmu@gmail.com}
\affiliation{
  \institution{Key Laboratory of Multimedia Trusted Perception and Efficient Computing,
  Ministry of Education of China,
  Xiamen University}
  \city{}
  \state{}
  \country{}
}

\author{Xiaoshuai Sun}\authornote{Corresponding author.}
\email{xssun@xmu.edu.cn}
\affiliation{
  \institution{Key Laboratory of Multimedia Trusted Perception and Efficient Computing,
  Ministry of Education of China,
  Xiamen University}
  \city{}
  \state{}
  \country{}
}

\author{Rongsheng Zhang}
\email{zhangrongsheng@corp.netease.com}
\affiliation{%
  \institution{Fuxi AI Lab, Netease Inc.}
  \state{}
  \country{}
}

\author{Yiwei Ma}
\email{yiweima@stu.xmu.edu.cn}
\affiliation{
  \institution{Key Laboratory of Multimedia Trusted Perception and Efficient Computing,
  Ministry of Education of China,
  Xiamen University}
  \city{}
  \state{}
  \country{}
}

\author{Minda Zhao}
\email{zhaominda01@corp.netease.com}
\affiliation{%
  \institution{Fuxi AI Lab, Netease Inc.}
  \state{}
  \country{}
}

\author{Lincheng Li}
\email{lilincheng@corp.netease.com}
\affiliation{%
  \institution{Fuxi AI Lab, Netease Inc.}
  \state{}
  \country{}
}

\author{Zeng Zhao}
\email{zengzhao_wl@163.com}
\affiliation{%
  \institution{Fuxi AI Lab, Netease Inc.}
  \state{}
  \country{}
}

\author{Tangjie Lv}
\email{hzlvtangjie@corp.netease.com}
\affiliation{%
  \institution{Fuxi AI Lab, Netease Inc.}
  \state{}
  \country{}
}

\author{Rongrong Ji}
\email{rrji@xmu.edu.cn}
\affiliation{
  \institution{Key Laboratory of Multimedia Trusted Perception and Efficient Computing,
  Ministry of Education of China, Xiamen University}
  \city{}
  \state{}
  \country{}
}


\begin{abstract}
In recent years, 3D understanding has turned to 2D vision-language pre-trained models to overcome data scarcity challenges. However, existing methods simply transfer 2D alignment strategies, aligning 3D representations with single-view 2D images and coarse-grained parent category text. These approaches introduce information degradation and insufficient synergy issues, leading to performance loss. Information degradation arises from overlooking the fact that a 3D representation should be equivalent to a series of multi-view images and more fine-grained subcategory text. Insufficient synergy neglects the idea that a robust 3D representation should align with the joint vision-language space, rather than independently aligning with each modality. In this paper, we propose a multi-view joint modality modeling approach, termed JM3D, to obtain a unified representation for point cloud, text, and image. Specifically, a novel Structured Multimodal Organizer (SMO) is proposed to address the information degradation issue, which introduces contiguous multi-view images and hierarchical text to enrich the representation of vision and language modalities. A Joint Multi-modal Alignment (JMA) is designed to tackle the insufficient synergy problem, which models the joint modality by incorporating language knowledge into the visual modality. Extensive experiments on ModelNet40 and ScanObjectNN demonstrate the effectiveness of our proposed method, JM3D, which achieves state-of-the-art performance in zero-shot 3D classification. JM3D outperforms ULIP by approximately 4.3\% on PointMLP and achieves an improvement of up to 6.5\% accuracy on PointNet++ in top-1 accuracy for zero-shot 3D classification on ModelNet40. The source code and trained models for all our experiments are publicly available at \url{https://github.com/Mr-Neko/JM3D}.
\end{abstract}


\renewcommand{\shortauthors}{Haowei Wang, et al.}



\maketitle

\section{Introduction \label{sec:Introduction}}

\begin{figure}[t]
\centering
\includegraphics[width=1.0\columnwidth]{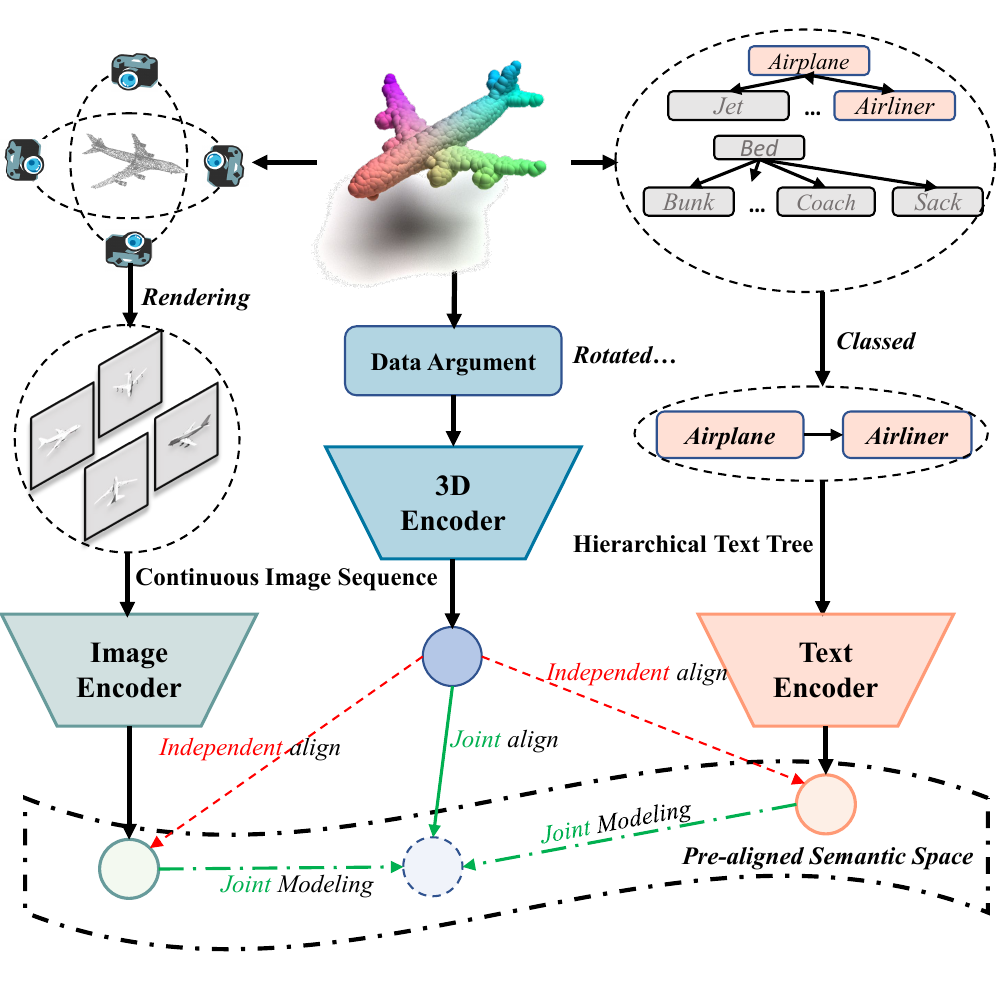}

\caption{The visualization of JM3D. JM3D aligns 3D modality with the pre-aligned vision and language modalities, constructing a unified representation of the three modalities. Continuous Image Sequence (CIS, left) and Hierarchical Text Tree (HTT, right) organize structured images and texts to enhance the information from vision and language modalities. The joint alignment and modeling {\color{green}(green line)} correct the inappropriate way of independent alignment {\color{red}(red line)} used in previous methods.}
\label{fig1}
\end{figure}


The representation learning of 3D models~\cite{achlioptas2018learning, liu2019densepoint, liu2020closer, ran2022surface, xie2020grnet, xu2021paconv} has become increasingly significant in various real-world applications such as augmented/virtual reality~\cite{liu2021group, vu2022softgroup, armeni20163d} and autonomous driving~\cite{li2022deepfusion, yin2021center}. Nevertheless, the development of 3D understanding is hindered by the scarcity of data and the inadequacy of category representation, contrasting sharply with the vast availability of image-text pairs.

To tackle the issue of insufficient 3D data, existing works~\cite{hegde2023clip, xue2022ulip, zhang2023clip} aim to harness the wealth of other modalities by employing large-scale vision-language models, such as CLIP~\cite{radford2021learning}, to improve 3D representation. The underlying principle of these approaches is to align the 3D features with the unified space of vision and language, thus benefiting from the powerful zero-shot capabilities of foundation models. These methods typically involve rendering an image of a 3D model from a specific angle, accompanied by a simple category label, and feeding them into CLIP. The 3D features are then aligned with the visual-language space via a contrastive approach. This strategy, which incorporates abundant external information, has been demonstrated to effectively enhance 3D understanding capabilities and exhibit good transferability, as evidenced by works such as ULIP~\cite{xue2022ulip} and CG3D~\cite{hegde2023clip}.

However, these methods predominantly adopt 2D alignment strategies for 3D representation learning, failing to account for the unique characteristics of 3D models. Consequently, they face two significant limitations: (1) \textbf{Information degradation}: the alignment of 3D representations with single-view images and coarse text leads to a loss of crucial spatial and depth information. For instance, when examining single-view images, the front render of an airplane in Fig.~\ref{fig1} lacks wing details, and so is the back render. Additionally, from a textual standpoint, the generic term ``airplane'' is insufficient to differentiate an airliner from other types of aircraft, such as jets or bombers. (2) \textbf{Insufficient synergy}:  these methods align 3D representations with image features and text features separately, neglecting the joint modeling of vision and language modalities. This issue complicates the optimization process for 3D representations, making it difficult to determine whether to move closer to image features or text features, and to what extent, ultimately leading to incomplete information utilization.

In response to the aforementioned limitations, we propose a more synergistic multi-modal approach for comprehensive 3D representation learning, named JM3D, as illustrated in Fig.~\ref{fig1}. The approach consists of two key components: the Structured Multi-modal Organizer (SMO) and the Joint Multi-modal Alignment (JMA). 
First, to mitigate information degradation, we design the SMO to enhance both vision and language modalities individually. For the vision, we believe that a 3D model should align with a continuous sequence of images from various viewpoints. Based on this premise, we introduce a Continuous Image Sequence (CIS) that jointly models a series of images. To further reduce information loss, we embed angle, color, and depth information into the images through encoding. For the language, we implement a Hierarchical Text Tree (HTT). By introducing sub-categories such as``jet'', ``airliner'', or ``bomber'', the model is encouraged to learn finer-grained information, thus enhancing generalizability. The inclusion of parent categories like "airplane" ensures that semantically similar sub-categories have similar features, thereby increasing the robustness of the representation.
Second, to address insufficient synergy, we propose the JMA method that jointly models visual and language modalities, considering both modalities simultaneously during optimization rather than in isolation. This approach allows the 3D representation to effectively capture comprehensive information from both vision and language modalities. More importantly, we systematically derive a theoretical framework to validate the effectiveness of this modeling approach, which can serve as a foundation for future related research. Extensive experiments demonstrate that our approach significantly improves 3D representation in the zero-shot 3D classification of ModelNet40 and ScanObjectNN, with respective performance gains of {\color{black}4.3}\% and {\color{black}6.5}\% for PointMLP and PointNet++ on the top-1 accuracy of the ModelNet40.

In summary, this paper presents three key contributions:
\begin{itemize}
    \item To address the information degradation issue, we propose the Structured Multimodal Organizer (SMO) that constructs a continuous multi-view sequence of rendered images and a hierarchical text tree. By enhancing visual and textual features, SMO compensates for the loss of 3D visual features, ensuring more comprehensive representations. 
    
    \item To tackle the insufficient synergy problem, we design the Joint Multi-modal Alignment (JMA), which incorporates both textual and visual modalities to obtain a joint representation. This approach significantly avoids suboptimal performance and facilitates a more cohesive understanding of the image-text pairs. 
    
    \item Our proposed approach, JM3D, achieves state-of-the-art performance on various downstream tasks, particularly in zero-shot 3D classification. On the ModelNet40 dataset, JM3D delivers a 4.3\% improvement for PointMLP and a 6.5\% improvement for PointNet++ on the top-1 accuracy of the ``All'' set of ModelNet40 compared to existing methods. 
\end{itemize}

\begin{figure*}[]
\centering
\includegraphics[width=1.0\textwidth]{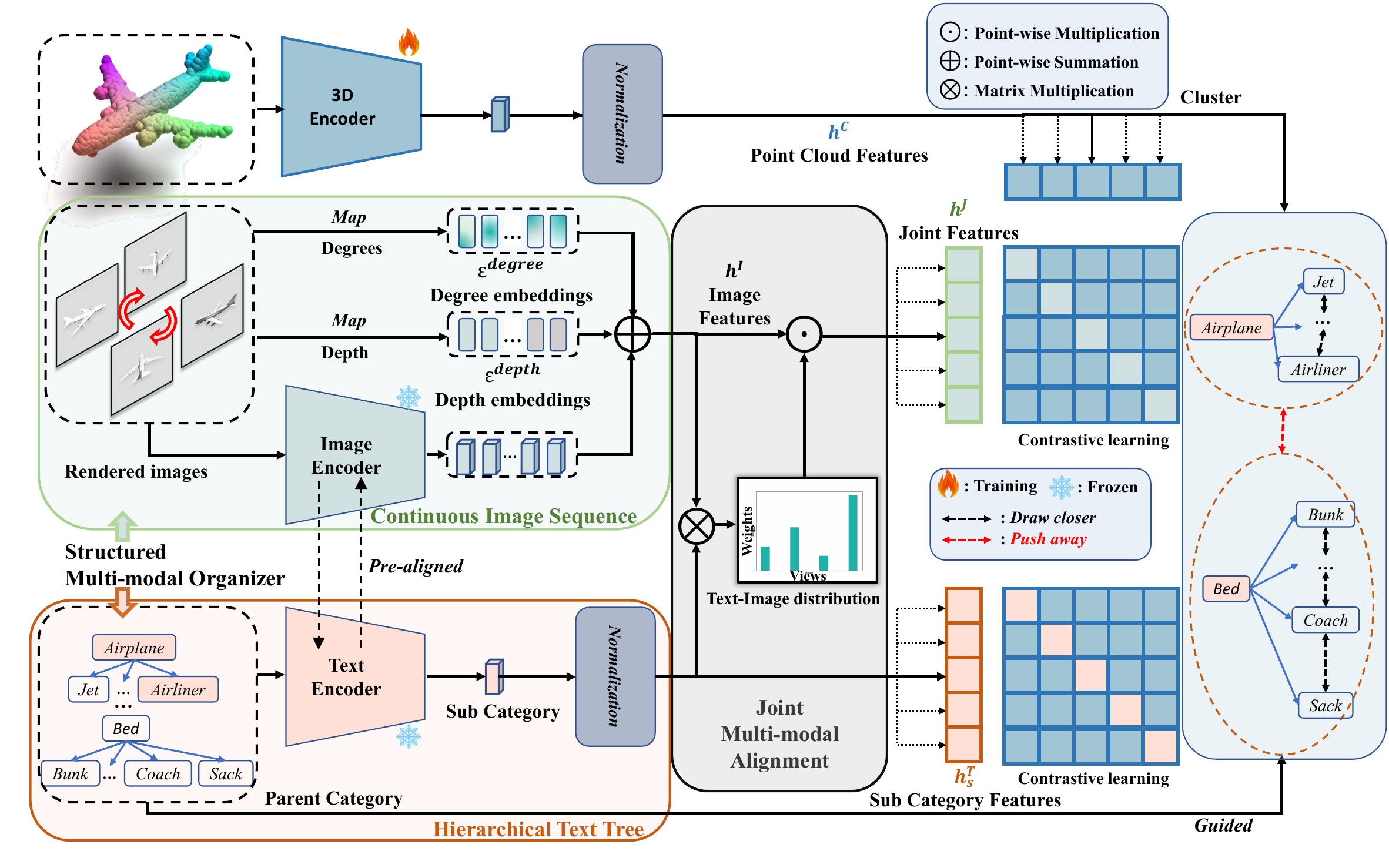}

\vspace{-0.25cm}
\caption{The framework of JM3D. Continuous Image Sequence {\color{green}(CIS)} and Hierarchical Text Tree {\color{orange}(HTT)} organized continuous multi-view images and hierarchical texts respectively, which are fed into a pre-training model (frozen) to extract features on the left. Then, Joint Multi-modal Alignment {\color{gray}(JMA)} incorporates the features from two modalities to generate the joint modeling features. On the last, contrastive learning is applied to align 3D features (training) with joint features and subcategory texts, while 3D features are aggregated with the assistance of the parent category.}
\label{fig2}
\end{figure*}

\section{Related work}
\subsection{Representation Learning in 3D Space}
The representation learning of 3D space is aimed to obtain semantic features of a 3D model. 3D models have different representation methods, among which point cloud has become one of the most suitable input formats in deep learning due to its sparsity and discreteness~\cite{aubry2011wave, bronstein2010scale, sun2009concise, wu20153d, maturana2015voxnet, zhao-etal-2023-generating-visual}. Early methods~\cite{maturana2015voxnet, shi2020pv} often extracted point cloud data from voxels and used some convolutions to extract global features. Subsequent methods attempted to design specific structures directly for point cloud, \emph{e.g.}, PointNet~\cite{qi2017pointnet}, PointNext~\cite{qian2022pointnext} and PointMLP~\cite{marethinking}. PointNet directly extracts permutation-invariant features from a point cloud, greatly influencing subsequent method designs. Recent PointMLP takes two MLP blocks and a geometrical transformer to achieve impressive results without integrating sophisticated local geometrical extractors.

With the rise of transformers~\cite{vaswani2017attention}, some self-supervised learning methods~\cite{guo2021pct, liu2022masked, xiao2023unsupervised} attempt to generate more point clouds and increase the difficulty of training tasks by using encoder-decoder structures to reconstruct point clouds, which is considered effective in such PointBert~\cite{yu2022point}, Point-MAE~\cite{pang2022masked}, and Point-M2AE~\cite{zhangpoint}. 

Even so, regardless of the structural design, the biggest challenge in 3D representation learning is the limited-scale dataset. Those methods are constrained by simple category annotations and insufficient data volume, resulting in mediocre robustness in the real world.

\subsection{Representation Learning in Multi-modal Space}
Representation learning in multi-modal space mainly aims at aligning semantics features from different modalities. Current methods primarily achieve modality alignment through two lines. One of them fused features from different modalities by delicately designed architectures~\cite{li2019visualbert, li2020oscar, chen2020uniter, hu2023you, fei-etal-2023-scene} and trained alignment through tasks such as retrieval and comprehension. These methods~\cite{li2019visualbert, li2020oscar, lu2019vilbert, tan2019lxmert, wang2023towards, jin2023refclip} often focus on learning the interaction between image regions and their corresponding descriptions.

Another line, such as CLIP~\cite{radford2021learning}, first applies contrastive learning on vision and language modalities, directly aligning features across different modalities with massive positive and negative samples. Some subsequent methods optimized the alignment of CLIP from both data scale and granularity perspectives. The series of GLIP~\cite{li2022grounded, zhang2022glipv2} attempted to achieve finer alignment by using detection tasks, while Flamingo~\cite{alayrac2022flamingo} introduced a scaled dataset. At the same time, approachs~\cite{luo2022clip4clip, li2022align, xu2021videoclip, ju2022prompting, FeiMatchStruICML22, ma2022xclip} have shown that the CLIP-paradigm remains effective even when extended to other modalities like video and text.
 
\subsection{Learning 3D Representation with Multi-modal}
Representation learning in multi-modal domains has shown that learning from different modalities brings better performance. Therefore, utilizing the rich knowledge of image-text pre-trained models to enhance the feature representation of 3D models is a promising way~\cite{chen2021multimodal, yanlet, ma2023xmesh}. PointCLIP~\cite{zhang2022pointclip} is the first attempt to utilize a visual-language pre-training model for point cloud models. It renders point cloud models as a series of depth images and directly feeds them into CLIP for zero-shot classification. However, it did not improve the expressive power of point clouds. ULIP~\cite{xue2022ulip} and CG3D~\cite{hegde2023clip} attempt to directly transplant the paradigm of CLIP to learn a unified representation space for point clouds, language, and images. Contrastive learning paradigm is operated between point cloud modality and visual modality as well as language modality, separately. These methods overlook the fact that the information contained in a single image or a piece of text cannot be equivalent to that of the entire 3D model. At the same time, aligning them separately ignores the joint distribution of image-text modalities. Our methods, on the one hand, with SMO reorganized the visual and language data, and the representations of the visual and language modality are strengthened. On the other hand, JMA modeled the joint features of the visual-language modality, optimizing the process of learning a unified semantic space.

\section{Method}

In this section, we will first review the contrastive learning framework for 3D representations in Sec.~\ref{sec:ulip}.  Next, the details of our proposed SMO module will be introduced in Sec.~\ref{sec:SMO}. Then we present the
JMA module in Sec.~\ref{sec:JMA}.  Finally, we present the objectives of the construction of unified triplet modality modeling in Sec.~\ref{sec:loss}. The overall framework is illustrated in Fig.~\ref{fig2}.

\subsection{Preliminary \label{sec:ulip}}

To acquire a unified semantic space across vision, language, and point cloud, ULIP~\cite{xue2022ulip} investigates the feasibility of directly transferring 2D contrastive learning to 3D. A dataset comprising point clouds, images, and textual descriptions is constructed from the ShapeNet55~\cite{chang2015shapenet}, which consists of publicly available 3D CAD models and their associated textual descriptions.  For the $i$-th CAD model, a triplet sample $S_i: (I_i, T_i, C_i)$ is created, consisting of an image $I_i \in \mathbb{R}^{H \times W \times 3}$, a text description $T_i$, and a sampled point cloud $C_i \in \mathbb{R}^{N_c}$, where $H$, $W$, and $N_c$ represent the height, width of the image, and the number of sampled points, respectively.
The image $I_i$ is rendered from the CAD model with a random angle, while the text description $T_i$ is created by concatenating a fixed prompt with a coarse pre-defined category, such as "a point cloud of [CLASS]". Meanwhile, the point cloud $C_i$ is uniformly sampled from the original models to accommodate different point cloud backbones. With these triplet samples, the representations of the three modalities are expected to align within a unified semantic space. The formulation of this alignment can be expressed as:
\begin{equation}
    P(C, I, T)=P(C|I, T) \cdot P(I, T).
    \label{eq:definition}
\end{equation}
Prior work~\cite{xue2022ulip, hegde2023clip} simplified Eq.~\ref{eq:definition} based on an approximate assumption, which states that the \textit{\textbf{2D image $I$ and text $T$ are conditionally independent}}. Consequently, ULIP does not consider the joint modality conditional probability $P(C|I, T)$, and instead simplifies it by aligning individual modalities as follows:
\begin{equation}
    P(C, I, T)=P(C|I) \cdot P(C|T) \cdot P(I, T).
    \label{eq:ULIP}
\end{equation}
Inspired by Eq.\ref{eq:ULIP}, ULIP learns the joint distribution contrastively, aligning the 3D features with language and vision features separately. Specifically, $P(I, T)$ can be represented by a pre-trained vision-language model, such as CLIP\cite{radford2021learning}. This pre-aligned model is employed to extract language and vision features, while 3D features can be extracted using various backbones~\cite{marethinking, qi2017pointnet++, yu2022point}. These operations can be formulated as:
\begin{equation}
    h_i^{C},\ h_i^I,\ h_i^T=f_{C}({C}_i),\ f_I(I_i),\ f_T(T_i),
    \label{eq:extract}
\end{equation}
where $f_{C}$ is a 3D representation network with various backbones. $f_I$ and $f_T$ are the pre-aligned vision encoder and language encoder from CLIP, which utilize vanilla transformer~\cite{vaswani2017attention} structures. The vectors $h_i^C\in \mathbb{R}^{D}$, $h_i^I\in \mathbb{R}^{D}$, and $h_i^T\in \mathbb{R}^{D}$ denote 3D, language, and vision features, respectively, and $D$ represents the dimensions of the final representation vectors.

Next, a contrastive method~\cite{radford2021learning} is chosen to model the conditional distribution between any two modalities in Eq.~\ref{eq:ULIP}, aligning $h_i^C$, $h_i^I$, and $h_i^T$ by:
\begin{equation}
 \mathcal{L}_{(h^{M_1}, h^{M_2})} =
   \sum_i-\frac{1}{2}log\frac{\exp\left({\mathbf{h}^{M_1}_i \mathbf{h}^{M_2}_i}\right)}{\sum_k \exp\left({\mathbf{h}^{M_1}_i \mathbf{h}^{M_2}_k}\right)},
    \label{eq:ULIP contrastive}
\end{equation}
where $M = (M_1, M_2) \in \{(T, I), (C, T), (C, I)\}$ represents the combination of pairwise modalities. 

\subsection{Structured Multi-modal Organizer \label{sec:SMO}}

To address the issue of information loss resulting from naively aligning 3D models $C_i$ with 2D vision and coarse-grained categories, we propose a more refined approach. For example, a car exists in 3D space, but a single frontal render lacks information about the rear end, as do other individual view renders. Similarly, the term "bottle" does not accurately represent specific models like ``jug'', ``beer'', ``bottle'' or ``flask''. To mitigate this information loss, we organize data in the triplet sample $S_i$ using a multi-view approach, constructing a new triplet data and redefining it as:
\begin{equation}
    S_i:\left(\left[I_{i1}, \cdots, I_{iv}\right], \left[T_{i}^p, T_{i}^s\right], C_i\right).
    \label{eq:sample}
\end{equation}
This refined representation leverages the $v$ images from Continuous Image Sequence (CIS) and structured texts with parent category and sub-category from Hierarchical Text Tree (HTT) to ensure a more accurate and detailed alignment between the 3D models and their corresponding 2D vision and textual descriptions.

\subsubsection{\textbf{Continuous Image Sequence}}
For the visual modality, single synthetic images from random angles only capture partial features of a 3D model. To enhance vision semantics, it is logical to introduce multi-view rendered images.
Building upon ULIP, we synthesize RGB and depth images at 12-degree intervals, creating a candidate image set $C_I \in \mathbb{R}^{N \times 3}$, where $N$ denotes the number of rendered images. Since the sampled images are discrete, we observed that large angular deviations between images in the sequence may cause training instability. To address this, we sample $v$ images within a specific angular range during each training process, as follows:
\begin{equation}
\left[I_{1}, \cdots, I_{v}\right] = WS(C_I), |\angle I_i-\angle I_j|<\omega, \forall i, j \in [1, v],
\end{equation}
where $WS(\cdot)$ signifies sampling within a specific angular range, $\angle I_i$ denotes the render degree of the $i$-th image, and $\omega$ is a hyperparameter set to $60^\circ$ based on our experiments.

The image encoder from pre-trained CLIP embeds multi-view images into the feature space. Furthermore, given the importance of angle and depth as positional information in images, we design angle and depth encodings, akin to \cite{vaswani2017attention}, to capture the 3D spatial information of different images. As illustrated in Fig.\ref{fig2}, these embeddings are added to the visual features, resulting in the following formulation:
\begin{equation}
h_{iv}^{I}=\textit{LayerNorm}\left(f_I(I_{iv})+\epsilon^{degree}[\angle I_i]+\epsilon^{depth}[\angle I_i]\right),
\end{equation}
where $\textit{LayerNorm}(\cdot)$, introduced in~\cite{1607.06450}, controls the range of vision vectors $h_{iv}^{I}$, consequently expediting convergence.

\subsubsection{\textbf{Hierarchical Text Tree}}
For the language modality, existing methods like ULIP use only simple parent categories (55 categories) for 3D representation learning. This approach reduces the number of categories to be learned, making the model easier to fit but also lowering its generalization performance. The ShapeNet55 dataset~\cite{chang2015shapenet} also provides subcategories (205 categories). Introducing subcategories forces the model to focus on finer-grained representations, enabling it to correctly distinguish between visually similar subcategories. However, it is important to note that when the dataset size is limited, independently introducing subcategories may cause the model to overlook the modeling of family relationships, where subcategories belonging to the same parent category should have more similar features. Incorporating these relationships greatly reduces the model's dependency on data size and improves convergence speed.

To address these issues, HTT constructs a novel hierarchical category tree for each point cloud model, comprising coarse semantic parent categories and more specific subcategories (e.g., ``bed'' $\rightarrow$ ``bunk''). Specifically, when subcategory annotations are partially missing for a model, we use its parent category as a replacement. Through this approach, HTT assigns structured category information to each model as $[T^p, T^s]$, where $T^p$ represents the parent category with primary semantics, and $T^s$ denotes the subcategory with fine-grained details.
With hierarchical-grained categories, we design specific tasks for both parent and subcategories. The subcategories $T^s$ follow Eq.~\ref{eq:extract} to generate text features using the textual encoder of CLIP, which will be learned in a contrastive manner. Parent categories are used to guide the aggregation of point cloud features, as expressed in the following formula:
\begin{equation}
\mathop{\arg\min}\limits_{\theta} g(\theta)={||\theta(h_{i}^{C})-f(T_{i}^{P})||}^2,
\label{eq:parent classed}
\end{equation}
where $\theta$ is an MLP~\cite{tolstikhin2021mlp} network, and $f(\cdot)$ maps the parent category of $i$-th sample to a fixed label. By performing this operation, the model can learn fine-grained features within limited samples while also possessing the ability to abstract family features, thus avoiding underfitting.

\subsection{Joint Multi-modal Alignment \label{sec:JMA}}
In previous methods, vision and language modalities were assumed to be conditionally independent, as demonstrated in Eq.\ref{eq:ULIP}. This approximation results in an \emph{insufficient synergy} issue when compared to Eq.\ref{eq:definition}, leading to suboptimal performance. To tackle this problem, we propose a more general approach by directly modeling the relationships between vision and language modalities, denoted as $P(C | I, T)$. This relationship can be reformulated as:
\begin{equation}
\begin{aligned}
    P(C|I, T)&=\frac{P(C, I|T)}{P(I|T)}=\frac{P(C, I|T)P(T)}{P(I|T)P(T)}=\frac{P(C, I|T)P(T)}{P(I, T)} \\
    &\propto \sum_{i}P(C, I|T_{i})P(T_{i})=\sum_{i}\sum_{j}P(C, I_{i, j}|T_{i})P(T_{i}),
    \label{eq:joint}
\end{aligned}
\end{equation}
which means that for any sample $S_i$ in Eq.~\ref{eq:sample}, we use both textual and multi-view visual information interactively for joint modeling. In practice, JMA generates multiple image features $h_{iv}^{I} \in \mathbb{R}^{V \times D}$ and fine-grained text feature $h_{is}^{T}\in \mathbb{R}^{D}$. Using these features, attention between text and all views is calculated to help reconstruct image features into joint modality features, as follows:
\begin{equation}
    h_i^J=\sum_v^V\text{Softmax}(h_{iv}^{I} \times h_{is}^{T}) \otimes h_{iv}^{I},
\end{equation}
where $\times$ denotes matrix multiplication, and $\otimes$ represents element-wise multiplication.

By leveraging JMA, we jointly map 2D image and textual information into a space that is more conducive to 3D features, obtaining a robust joint image-text representation feature. This enhanced feature further aids in aligning the 3D modality with the other two modalities using the contrastive learning paradigm, ultimately resulting in improved performance.

\subsection{Training Objective \label{sec:loss}}

For the training process, we set two tasks for the unified 3D representation learning. First, a contrastive way is set between the point cloud representation, language features like Eq.~\ref{eq:ULIP contrastive}, which is formulated as:
\begin{equation}
    \mathcal{L}_{contrastive}=\lambda_1\mathcal{L}_{(h^{C}, h^J)} + \lambda_2\mathcal{L}_{(h^{C}, h_s^T)} + \lambda_3\mathcal{L}_{(h_s^T, h^J)},
\end{equation}
where the $\lambda_1$, $\lambda_2$, and $\lambda_3$ are the hyper parameters. The contrastive learning draws the features across three modalities to a unified semantic space, while the $h_s^T$ from the subcategory provides guidance in fine-grained. 

At the same time, another task is the classification task constructed for the parent category as Eq.~\ref{eq:parent classed}, which is formulated as:

\begin{equation}
    \mathcal{L}_{classed}=\frac{1}{N}\sum_{i}\sum_{j}^{|T^p|}f(T_{i, j}^{p})log(\text{Softmax}(\theta(h_{i}^{C}))),
\end{equation}
where the $\theta$ is the parameters of a MLP, and $f(T_{i, j}^{p})$ is the probability of $j$-th parent category for the $i$-th sample. In this manner, point cloud features are aggregated under the guidance of the parent category, providing a softer constraint compared to the strict contrastive learning approach. This simple method, through the introduction of parent categories, significantly reduces the difficulty in fitting subcategories. The final loss is obtained by summing the two aforementioned loss. 

\begin{table*}[htb]
    \caption{The results of zero-shot 3D classification on ModelNet40 and ScanObjectNN datasets. PointMLP + JM3D outperforms the previous state-of-the-art methods by a large margin in various evaluation settings, especially achieving a 12.3\% and 13.7\% improvement of the ``Medium'' and ``Hard'' mode on ModelNet40, which is the SOTA.}
    \centering
    \begin{tabular}{lcccccccc}
         \toprule
         \multirow{3}*{Model} & \multicolumn{6}{c}{ModelNet40} & \multicolumn{2}{c}{ScanObjectNN} \\
         
         \cmidrule(lr){2-7}\cmidrule(lr){8-9}
         
         ~ & \multicolumn{2}{c}{ All } & \multicolumn{2}{c}{Medium} & \multicolumn{2}{c}{Hard} & \multicolumn{2}{c}{All}
         \\
         \cmidrule(lr){2-3}\cmidrule(lr){4-5}\cmidrule(lr){6-7}\cmidrule(lr){8-9}
         ~  & top-1 & top-5 & top-1 & top-5 & top-1 & top-5 & top-1 & top-5
         \\
         \midrule
         PointCLIP~\cite{zhang2022pointclip}  & 20.2 & -- & 10.4 & --& 8.3 &-- & 15.4 & --\\
         PointMLP~\cite{marethinking} + CG3D~\cite{hegde2023clip} & 50.4 & -- & -- & -- & -- & -- & 25.0 & --\\
         PointTransformer~\cite{zhao2021point} + CG3D~\cite{hegde2023clip} & 50.6 & -- & -- & -- & -- & -- & 25.6 & --\\
         PointNet++(ssg)~\cite{qi2017pointnet++} + ULIP~\cite{xue2022ulip}& 55.7 & 75.7 & 35.6 & 64.4 & 33.7 & 55.8 & 45.6 & 73.8\\
         PointBERT~\cite{yu2022point} + ULIP~\cite{xue2022ulip}& 60.4 & \textbf{84.0} & 40.4 & 72.1 & 37.1 & 66.3 & 48.5 & 79.9\\
         PointMLP~\cite{marethinking} + ULIP~\cite{xue2022ulip} & 61.5 & 80.7 & 43.2 & 72.0 & 36.3 & 65.0 & 44.6 & 82.3\\
         \midrule
         PointNet++(ssg)~\cite{qi2017pointnet++} + JM3D & 62.2 \textcolor{darkgreen}{\small ($\uparrow 6.5$)} & 79.3 & 47.6 \textcolor{darkgreen}{\small ($\uparrow 12.0$)} & 75.9 & 43.3 \textcolor{darkgreen}{\small ($\uparrow 9.6$)} & 74.7 & 46.0 \textcolor{darkgreen}{\small ($\uparrow 0.4$)} & 78.1\\
         
         PointBERT~\cite{yu2022point} + JM3D & 61.8 \textcolor{darkgreen}{\small ($\uparrow 1.4$)} & 81.7 & 52.9 \textcolor{darkgreen}{\small ($\uparrow 12.5$)} & 73.6 & 48.4 \textcolor{darkgreen}{\small ($\uparrow 11.3$)} & 71.1 & \textbf{48.9} \textcolor{darkgreen}{\small ($\uparrow 0.4$)} & 82.8\\
         
         PointMLP~\cite{marethinking} + JM3D & \textbf{65.8} \textcolor{darkgreen}{\small ($\uparrow 4.3$)} & 82.1 & \textbf{55.5} \textcolor{darkgreen}{\small ($\uparrow 12.3$)} & \textbf{77.1} & \textbf{55.0} \textcolor{darkgreen}{\small ($\uparrow 13.7$)} & \textbf{75.0} & 47.5 \textcolor{darkgreen}{\small ($\uparrow 2.9$)} & \textbf{83.3}\\

         \bottomrule
    \end{tabular}
    \label{tab:zero-shot-modelnet}
\end{table*}

\begin{table*}[htb]
    \caption{The ablation study of CIS. Multi-views will lead to point cloud features being biased towards vision modality, which decreases the performance on zero-shot 3D classification. However, CIS effectively improves the performance with the embeddings and within-view sample.}
    \centering
    \begin{tabular}{ccc|cccccccc}
         \toprule
         \multirow{3}*{Views} & \multirow{3}*{Sample} & \multirow{3}*{Embedding}  & \multicolumn{6}{c}{ModelNet40} & \multicolumn{2}{c}{ScanObjectNN} \\
         
         \cmidrule(lr){4-9}\cmidrule(lr){10-11}
         
         ~ & ~ & ~ & \multicolumn{2}{c}{ All } & \multicolumn{2}{c}{Medium} & \multicolumn{2}{c}{Hard} & \multicolumn{2}{c}{All}
         \\
         \cmidrule(lr){4-5}\cmidrule(lr){6-7}\cmidrule(lr){8-9}\cmidrule(lr){10-11}
         ~ & ~ & ~ & top-1 & top-5 & top-1 & top-5 & top-1 & top-5 & top-1 & top-5 \\
         \midrule
         1 & random & \ding{55} & 60.0 & 79.4 & 43.2 & 72.0 & 36.3 & 65.0 & 44.6 & 82.3\\
         4 & random & \ding{55} & 56.5 & 76.6 & 42.3 & 70.4 & 38.2 & 66.3 & 44.5 & 75.3\\
         4 & random & \ding{51} & 58.8 & 79.9 & 47.6 & \textbf{75.1} & 47.5 & 71.2 & 44.9 & 81.2\\
         4 & within-view & \ding{51} & 60.7 & 80.2 & 48.7 & 74.7 & 50.0 & 71.2 & \textbf{45.4} & \textbf{82.8}\\
         2 & within-view & \ding{51} & \textbf{61.2} & \textbf{80.7} & \textbf{49.5} & 74.5 & \textbf{51.4} & \textbf{71.7} & \textbf{45.4} & 82.2\\

         \bottomrule
    \end{tabular}
    \label{tab:ablation-image}
\end{table*}

\begin{table*}[htb]
    \caption{The ablation study of HTT. Results show that structured text is more effective than fine-grained text. Even if the language modality is enhanced, the independent alignment method still makes the improvements unstable.}
    \centering
    \begin{tabular}{cc|cccccccc}
         \toprule
         \multirow{3}*{Contrastive Text} & \multirow{3}*{Classed Text}  & \multicolumn{6}{c}{ModelNet40} & \multicolumn{2}{c}{ScanObjectNN} \\
         
         \cmidrule(lr){3-8}\cmidrule(lr){9-10}
         
         ~ & ~ & \multicolumn{2}{c}{ All } & \multicolumn{2}{c}{Medium} & \multicolumn{2}{c}{Hard} & \multicolumn{2}{c}{All}
         \\
         \cmidrule(lr){3-4}\cmidrule(lr){5-6}\cmidrule(lr){7-8}\cmidrule(lr){9-10}
         ~ & ~ & top-1 & top-5 & top-1 & top-5 & top-1 & top-5 & top-1 & top-5 \\
         \midrule
         Parent category & \ding{55} & 61.2& 80.7 & 49.5 & 74.5 & \textbf{51.4} & 71.7 & 45.4 & \textbf{82.2}\\
         Subcategory & \ding{55} & 61.8 & 81.0 & 48.5 & \textbf{78.6} & 43.5 & \textbf{75.2} & 46.4 & 79.5\\
         Subcategory & Parent category & \textbf{63.1} & \textbf{81.4} & \textbf{49.7} & 76.7 & 50.4 & 74.4 & \textbf{46.9} & 80.4\\

         \bottomrule
    \end{tabular}
    \label{tab:ablation-text}
    \vspace{-0.1cm}
\end{table*}

\begin{table*}[htb]
    \caption{The ablation study for JMA. Independent alignment wastes the rich semantics brought by SMO, while JMA achieves significant improvements on all settings of different datasets.}
    \centering
    \begin{tabular}{ccc|cccccccc}
         \toprule
         \multirow{3}*{CIS} & \multirow{3}*{HTT} & \multirow{3}*{JMA}  & \multicolumn{6}{c}{ModelNet40} & \multicolumn{2}{c}{ScanObjectNN} \\
         
         \cmidrule(lr){4-9}\cmidrule(lr){10-11}
         
         ~ & ~ & ~ & \multicolumn{2}{c}{ All } & \multicolumn{2}{c}{Medium} & \multicolumn{2}{c}{Hard} & \multicolumn{2}{c}{All}
         \\
         \cmidrule(lr){4-5}\cmidrule(lr){6-7}\cmidrule(lr){8-9}\cmidrule(lr){10-11}
         ~ & ~ & ~ & top-1 & top-5 & top-1 & top-5 & top-1 & top-5 & top-1 & top-5 \\
         \midrule
         \ding{55} & \ding{55} & \ding{55} & 60.0 & 79.4 & 43.2 & 72.0 & 36.3 & 65.0 & 44.6 & 82.3\\
         \ding{51} & \ding{51} & \ding{55} & 63.1 & 81.4 & 49.7 & 76.7 & 50.4 & 74.4 & 46.9 & 80.4\\
         \ding{51} & \ding{51} & \ding{51} & \textbf{65.8} & \textbf{82.1} & \textbf{55.5} & \textbf{77.1} & \textbf{55.0} & \textbf{75.0} & \textbf{47.5} & \textbf{83.3}\\

         \bottomrule
    \end{tabular}
    \label{tab:ablation-joint}
\end{table*}

\section{Experiment}
\subsection{Datasets}
\subsubsection{\textbf{Pretrain Datasets}}
We use \textbf{ShapeNet55}~\cite{chang2015shapenet} as the pre-training dataset, which is the publicly-available subset of ShapeNet. ShapeNet consists of 52.5K CAD models with multiple texture maps and corresponding category annotations. The annotations have a total of 55 basic categories and 205 fine-grained subcategories, with a small number of models missing subcategories. During training, we randomly sample different numbers of points from the CAD models to adapt different networks.
\subsubsection{\textbf{Downstream Datasets}}
We conducted downstream task experiments primarily on the following two datasets.

\textbf{ModelNet40}~\cite{wu20153d} is composed of synthetic 3D CAD models, consisting of 9,843 training samples and 2,468 testing samples, spanning across 40 categories. For the test, we follow~\cite{marethinking} to downsample the point cloud data to 1024 points.

\textbf{ScanObjectNN}~\cite{uy2019revisiting} is different from ModelNet40. It is a dataset of scanned 3D objects from real scenes, which consists of 2,902 samples among 15 categories. It can be divided into two categories based on whether or not it includes background: \emph{OBJ\_ONLY} and \emph{OBJ\_BJ}. The former refers to a clean mesh, while the latter includes background noise. Here we follow ULIP to use the pre-processed data~\cite{uy2019revisiting} from~\cite{yu2022point}, which is normalized and downsampled to 1024 points.

\subsection{3D Backbone Networks}
To verify the effectiveness of the proposed JM3D, we conducted experiments on ModelNet40 and ScanObjectNN with different 3D backbones, which are:

\textbf{PointNet++}~\cite{qi2017pointnet++} is the sequel of PointNet~\cite{qi2017pointnet}, which is an encoder-decoder formulation to extract the deep hierarchical features on point sets. The encoder is composed of many set abstraction modules, and the farthest point sample is between them to reduce the scale of point sets. For the decoder, the outputs of all the encoder layers will be sent to different head networks to adapt to diverse tasks.

\textbf{PointMLP}~\cite{marethinking} is a lite network with residual MLP modules to better extract features, avoiding the complicated architecture to utilize the local geometric features. PointMLP introduces Geometric Affine Module to transfer points to the normal distribution, and two MLP blocks are used to extract the representation and position information respectively. 

\textbf{PointBert}~\cite{yu2022point} is a transformer-based model to introduce self-supervised learning to the the point cloud representation learning. By reconstructing the masked point cloud, it achieves impressive performances in unlabeled point cloud datasets. PointBert randomly masks a number of points and learns stability features by reconstructing the masked points. 

\subsection{Implementation Details}
\subsubsection{\textbf{Pre-training}}

We uniformly sample the point cloud to 1024, 2048, and 8192 points to match the recommended setting of different backbones. Rendered images and texts are pre-processed as the pre-trained image-text encoder requirements. We use SLIP~\cite{mu2022slip} instead of the original CLIP model to get better performance, while the image and text encoder in our experiment is frozen, just like ULIP. During training, only the parameters in point cloud backbone is trainable. We train JM3D for 250 epochs. The batch size is 128 with a learning rate of $1e-3$. And AdamW is the optimizer, while the Cosine LR schedule is utilized.

\subsubsection{\textbf{Zero-shot 3D Classification}}

JM3D measures the distance between the point cloud features and the text features from new datasets, while the category with the smallest distance is selected. The pre-processing of text and point cloud is the same as pre-training, and there is no finetuning stage involved. We conduct zero-shot evaluations on both ModelNet40 and ScanObjectNN. The former dataset consists of a synthetic model and unseen categories, the results of which can indicate the alignment effects of multi-modal features. ScanObjectNN is a challenged dataset with real-world scanned data, which challenges the robustness of 3D pre-trained models.



\subsection{Zero-shot 3D Classification}


JM3D achieves zero-shot 3D recognition by computing the similarity between point cloud features and textural features. In this section, we present experimental results that demonstrate the cross-modal understanding capabilities of JM3D by conducting zero-shot experiments on two different evaluation sets. We use the top-1 accuracy as the main metric, which indicates the ability of aligned features in the sense of practical application. 

\subsubsection{\textbf{Evaluation Sets.}} 
To be fair with the previous methods~\cite{xue2022ulip, zhang2022pointclip, hegde2023clip}, we conduct experiments on the ModelNet40 and ScanObjctNN datasets by a single model. The ``All'' column in Tab.~\ref{tab:zero-shot-modelnet} represents results on all test samples. Furthermore, in order to distinguish similar classes that appear in the ModelNet and pre-trained datasets, which clearly affect the fairness of the zero-shot evaluation, we separately create a ``Medium'' set and a ``Hard'' set for ModelNet40. The ``Medium'' set excludes the common classes between ModelNet40 and ShapeNet55, while the ``Hard'' set further excludes semantically similar classes, \emph{e.g.}, ``chair'' vs ``stool'', to ensure that all categories in this set have not been leaked. For each set, we compute top-1 accuracy and top-5 accuracy as evaluation metrics. Top-1 refers to the model finding the text with the smallest distance to the point cloud feature, and we believe that the top-1 metric is more intuitive in demonstrating the model's zero-shot ability.

\subsubsection{\textbf{Experiment Results.}}
We present the zero-shot results on ModelNet40 and ScanObjctNN in Tab.~\ref{tab:zero-shot-modelnet}. Firstly, our method outperforms previous SOTA ULIP~\cite{xue2022ulip} across all 3D backbones on the top-1 accuracy on all sets with varying degrees of improvement. Specifically, compared to the ULIP, our best method, JM3D + PointMLP, in the case of the same backbone, still improves the top-1 accuracy on the ``All'' set by 4.3\%, on ``Medium'' set by 12.3\%, and on ``Hard'' set by 13.7\%, which shows that superiority of JM3D. Then, we also demonstrate the effectiveness of JM3D on ScanObjectNN. It can be seen that our JM3D + PointMLP approach outperforms ULIP in terms of top-1 accuracy by 2.9\%. All in all, we take impressive progress beyond the previous SOTA method ULIP~\cite{xue2022ulip}. This indicates that JM3D has good generalization and performs better in actual scanning scenarios.




\subsection{Ablation Study}

To investigate the exact contributions of SMO and JMA to the pre-training process, we conduct independent ablation studies of the two modules on PointMLP. Considering that the goal of validation is the alignment of point cloud features and image-text features, we respectively use the zero-shot metrics on both ModelNet40 and ScanObjectNN and cross-modal retrieval as qualitative evaluation criteria.



\subsubsection{\textbf{Continuous Multi Views vs. Random One Look}}

The comparison result is illustrated in Tab.~\ref{tab:ablation-image}. First, we conduct an ablation study on the number of images to investigate the impact of different numbers of viewpoint images. We found that directly introducing multiple viewpoints leads to a falling on the performance. This is because it disrupts the semantic continuity of different viewpoint features in a crude manner, causing the model to be confused with the different information from multiple viewpoints. It can be seen that with the addition of embeddings, the model's ability to distinguish viewpoints is enhanced, resulting in a 2.3\% improvement. After adding within-view sampling, the semantic continuity of viewpoints is further ensured with a 1.9\% improvement. However, too many images push the model more likely to align with image features, which results in a performance drop when the number of images increases from 2 to 4. At the same time, the image retrieval ability improves as shown in Fig~\ref{fig3}, which will be further explained in Appendix.\ref{sec:cross-retrival}.


\subsubsection{\textbf{Hierarchical Text vs. Pre-difined Text}}

After demonstrating the effectiveness of CIS, we show the role of the text tree in Tab.~\ref{tab:ablation-text}. We can see that simply introducing subcategory text does not bring significant improvement, only leads to a increase of 0.6\% on top-1 accuracy. This indicates that the granularity of the text is not the key issue. However, after introducing the structured category tree, {i.e.}, HTT, a 1.3\% improvement is achieved. Meanwhile, we notice that the top-5 accuracy with subcategories decreases, which is because more categories increase the difficulty of aligning language features. In general, through HTT, it reduces the distance between samples within each parent category cluster. HTT brings structured semantic information, which improves the alignments between point cloud features and text features.


\subsubsection{\textbf{Collaborative Alignment vs. Independent Alignment}}

As shown on Tab.~\ref{tab:ablation-joint}, JMA brings a 2.7\% improvement in the top-1 accuracy, which demonstrates the effectiveness of the JMA in experience. Even with the high-quality data organization brought by SMO, the improvement in the model's performance is not significant. Due to the incorrect alignment methods, the alignment of point cloud features and the alignment of text-image features is unstable. Adding JMA allows us to extend the strong assumption of independent alignment with text-image modality to the joint modeling scenario, significantly improving the alignment of point clouds.

\section{Conclusion}

We propose JM3D, a comprehensive pre-training framework with SMO and JMA that seamlessly integrates language, image, and point cloud features into a unified semantic space without any specialized design. By meticulously organizing data, the SMO module fully leverages information from each modality and the JMA module pioneers joint modeling to optimize modality alignment. Ablation studies validate the efficacy of the proposed SMO and JMA. Additionally, JM3D's superior performance in zero-shot 3D classification and image retrieval tasks, setting a new state-of-the-art, highlights its exceptional cross-modal capabilities. In the future, we will explore diverse data and alternative joint modeling methods to further expand the unified representation learning in 3D.

\begin{acks}

This work was supported by National Key R\&D Program of China (No.2022ZD0118201), the National Science Fund for Distinguished Young Scholars (No.62025603), the National Natural Science Foundation of China (No. U21B2037, No. U22B2051, No. 62176222, No. 62176223, No. 62176226, No. 62072386, No. 62072387, No. 62072389, No. 62002305 and No. 62272401), China Postdoctoral Science Foundation (No.2023M732948), and the Natural Science Foundation of Fujian Province of China (No.2021J01002,  No.2022J06001).
\end{acks}

\bibliographystyle{ACM-Reference-Format}
\balance
\bibliography{acmart}


\begin{thebibliography}{66}


\ifx \showCODEN    \undefined \def \showCODEN     #1{\unskip}     \fi
\ifx \showDOI      \undefined \def \showDOI       #1{#1}\fi
\ifx \showISBNx    \undefined \def \showISBNx     #1{\unskip}     \fi
\ifx \showISBNxiii \undefined \def \showISBNxiii  #1{\unskip}     \fi
\ifx \showISSN     \undefined \def \showISSN      #1{\unskip}     \fi
\ifx \showLCCN     \undefined \def \showLCCN      #1{\unskip}     \fi
\ifx \shownote     \undefined \def \shownote      #1{#1}          \fi
\ifx \showarticletitle \undefined \def \showarticletitle #1{#1}   \fi
\ifx \showURL      \undefined \def \showURL       {\relax}        \fi
\providecommand\bibfield[2]{#2}
\providecommand\bibinfo[2]{#2}
\providecommand\natexlab[1]{#1}
\providecommand\showeprint[2][]{arXiv:#2}

\bibitem[Achlioptas et~al\mbox{.}(2018)]%
        {achlioptas2018learning}
\bibfield{author}{\bibinfo{person}{Panos Achlioptas}, \bibinfo{person}{Olga
  Diamanti}, \bibinfo{person}{Ioannis Mitliagkas}, {and}
  \bibinfo{person}{Leonidas Guibas}.} \bibinfo{year}{2018}\natexlab{}.
\newblock \showarticletitle{Learning representations and generative models for
  3d point clouds}. In \bibinfo{booktitle}{\emph{International conference on
  machine learning}}. PMLR, \bibinfo{pages}{40--49}.
\newblock


\bibitem[Alayrac et~al\mbox{.}(2022)]%
        {alayrac2022flamingo}
\bibfield{author}{\bibinfo{person}{Jean-Baptiste Alayrac},
  \bibinfo{person}{Jeff Donahue}, \bibinfo{person}{Pauline Luc},
  \bibinfo{person}{Antoine Miech}, \bibinfo{person}{Iain Barr},
  \bibinfo{person}{Yana Hasson}, \bibinfo{person}{Karel Lenc},
  \bibinfo{person}{Arthur Mensch}, \bibinfo{person}{Katherine Millican},
  \bibinfo{person}{Malcolm Reynolds}, {et~al\mbox{.}}}
  \bibinfo{year}{2022}\natexlab{}.
\newblock \showarticletitle{Flamingo: a visual language model for few-shot
  learning}.
\newblock \bibinfo{journal}{\emph{Advances in Neural Information Processing
  Systems}}  \bibinfo{volume}{35} (\bibinfo{year}{2022}),
  \bibinfo{pages}{23716--23736}.
\newblock


\bibitem[Armeni et~al\mbox{.}(2016)]%
        {armeni20163d}
\bibfield{author}{\bibinfo{person}{Iro Armeni}, \bibinfo{person}{Ozan Sener},
  \bibinfo{person}{Amir~R Zamir}, \bibinfo{person}{Helen Jiang},
  \bibinfo{person}{Ioannis Brilakis}, \bibinfo{person}{Martin Fischer}, {and}
  \bibinfo{person}{Silvio Savarese}.} \bibinfo{year}{2016}\natexlab{}.
\newblock \showarticletitle{3d semantic parsing of large-scale indoor spaces}.
  In \bibinfo{booktitle}{\emph{Proceedings of the IEEE conference on computer
  vision and pattern recognition}}. \bibinfo{pages}{1534--1543}.
\newblock


\bibitem[Aubry et~al\mbox{.}(2011)]%
        {aubry2011wave}
\bibfield{author}{\bibinfo{person}{Mathieu Aubry}, \bibinfo{person}{Ulrich
  Schlickewei}, {and} \bibinfo{person}{Daniel Cremers}.}
  \bibinfo{year}{2011}\natexlab{}.
\newblock \showarticletitle{The wave kernel signature: A quantum mechanical
  approach to shape analysis}. In \bibinfo{booktitle}{\emph{2011 IEEE
  international conference on computer vision workshops (ICCV workshops)}}.
  IEEE, \bibinfo{pages}{1626--1633}.
\newblock


\bibitem[Ba et~al\mbox{.}(2016)]%
        {1607.06450}
\bibfield{author}{\bibinfo{person}{Jimmy~Lei Ba}, \bibinfo{person}{Jamie~Ryan
  Kiros}, {and} \bibinfo{person}{Geoffrey~E. Hinton}.}
  \bibinfo{year}{2016}\natexlab{}.
\newblock \bibinfo{title}{Layer Normalization}.
\newblock
\newblock
\showeprint[arxiv]{1607.06450}


\bibitem[Bronstein and Kokkinos(2010)]%
        {bronstein2010scale}
\bibfield{author}{\bibinfo{person}{Michael~M Bronstein} {and}
  \bibinfo{person}{Iasonas Kokkinos}.} \bibinfo{year}{2010}\natexlab{}.
\newblock \showarticletitle{Scale-invariant heat kernel signatures for
  non-rigid shape recognition}. In \bibinfo{booktitle}{\emph{2010 IEEE computer
  society conference on computer vision and pattern recognition}}. IEEE,
  \bibinfo{pages}{1704--1711}.
\newblock


\bibitem[Chang et~al\mbox{.}(2015)]%
        {chang2015shapenet}
\bibfield{author}{\bibinfo{person}{Angel~X Chang}, \bibinfo{person}{Thomas
  Funkhouser}, \bibinfo{person}{Leonidas Guibas}, \bibinfo{person}{Pat
  Hanrahan}, \bibinfo{person}{Qixing Huang}, \bibinfo{person}{Zimo Li},
  \bibinfo{person}{Silvio Savarese}, \bibinfo{person}{Manolis Savva},
  \bibinfo{person}{Shuran Song}, \bibinfo{person}{Hao Su}, {et~al\mbox{.}}}
  \bibinfo{year}{2015}\natexlab{}.
\newblock \showarticletitle{Shapenet: An information-rich 3d model repository}.
\newblock \bibinfo{journal}{\emph{arXiv preprint arXiv:1512.03012}}
  (\bibinfo{year}{2015}).
\newblock


\bibitem[Chen et~al\mbox{.}(2020)]%
        {chen2020uniter}
\bibfield{author}{\bibinfo{person}{Yen-Chun Chen}, \bibinfo{person}{Linjie Li},
  \bibinfo{person}{Licheng Yu}, \bibinfo{person}{Ahmed El~Kholy},
  \bibinfo{person}{Faisal Ahmed}, \bibinfo{person}{Zhe Gan},
  \bibinfo{person}{Yu Cheng}, {and} \bibinfo{person}{Jingjing Liu}.}
  \bibinfo{year}{2020}\natexlab{}.
\newblock \showarticletitle{Uniter: Universal image-text representation
  learning}. In \bibinfo{booktitle}{\emph{Computer Vision--ECCV 2020: 16th
  European Conference, Glasgow, UK, August 23--28, 2020, Proceedings, Part
  XXX}}. Springer, \bibinfo{pages}{104--120}.
\newblock


\bibitem[Chen and Jing(2021)]%
        {chen2021multimodal}
\bibfield{author}{\bibinfo{person}{Z Chen} {and} \bibinfo{person}{L Jing}.}
  \bibinfo{year}{2021}\natexlab{}.
\newblock \showarticletitle{Multimodal Semi-Supervised Learning for 3D
  Objects}. In \bibinfo{booktitle}{\emph{The British Machine Vision Conference
  (BMVC)}}.
\newblock


\bibitem[Fei et~al\mbox{.}(2023)]%
        {fei-etal-2023-scene}
\bibfield{author}{\bibinfo{person}{Hao Fei}, \bibinfo{person}{Qian Liu},
  \bibinfo{person}{Meishan Zhang}, \bibinfo{person}{Min Zhang}, {and}
  \bibinfo{person}{Tat-Seng Chua}.} \bibinfo{year}{2023}\natexlab{}.
\newblock \showarticletitle{Scene Graph as Pivoting: Inference-time Image-free
  Unsupervised Multimodal Machine Translation with Visual Scene Hallucination}.
  In \bibinfo{booktitle}{\emph{Proceedings of the 61st Annual Meeting of the
  Association for Computational Linguistics (Volume 1: Long Papers)}}.
  \bibinfo{pages}{5980--5994}.
\newblock


\bibitem[Fei et~al\mbox{.}(2022)]%
        {FeiMatchStruICML22}
\bibfield{author}{\bibinfo{person}{Hao Fei}, \bibinfo{person}{Shengqiong Wu},
  \bibinfo{person}{Yafeng Ren}, {and} \bibinfo{person}{Meishan Zhang}.}
  \bibinfo{year}{2022}\natexlab{}.
\newblock \showarticletitle{Matching Structure for Dual Learning}. In
  \bibinfo{booktitle}{\emph{Proceedings of the International Conference on
  Machine Learning, {ICML}}}. \bibinfo{pages}{6373--6391}.
\newblock


\bibitem[Fei-Fei et~al\mbox{.}(2004)]%
        {fei2004learning}
\bibfield{author}{\bibinfo{person}{Li Fei-Fei}, \bibinfo{person}{Rob Fergus},
  {and} \bibinfo{person}{Pietro Perona}.} \bibinfo{year}{2004}\natexlab{}.
\newblock \showarticletitle{Learning generative visual models from few training
  examples: An incremental bayesian approach tested on 101 object categories}.
  In \bibinfo{booktitle}{\emph{2004 conference on computer vision and pattern
  recognition workshop}}. IEEE, \bibinfo{pages}{178--178}.
\newblock


\bibitem[Guo et~al\mbox{.}(2021)]%
        {guo2021pct}
\bibfield{author}{\bibinfo{person}{Meng-Hao Guo}, \bibinfo{person}{Jun-Xiong
  Cai}, \bibinfo{person}{Zheng-Ning Liu}, \bibinfo{person}{Tai-Jiang Mu},
  \bibinfo{person}{Ralph~R Martin}, {and} \bibinfo{person}{Shi-Min Hu}.}
  \bibinfo{year}{2021}\natexlab{}.
\newblock \showarticletitle{Pct: Point cloud transformer}.
\newblock \bibinfo{journal}{\emph{Computational Visual Media}}
  \bibinfo{volume}{7} (\bibinfo{year}{2021}), \bibinfo{pages}{187--199}.
\newblock


\bibitem[Hamdi et~al\mbox{.}(2021)]%
        {hamdi2021mvtn}
\bibfield{author}{\bibinfo{person}{Abdullah Hamdi}, \bibinfo{person}{Silvio
  Giancola}, {and} \bibinfo{person}{Bernard Ghanem}.}
  \bibinfo{year}{2021}\natexlab{}.
\newblock \showarticletitle{Mvtn: Multi-view transformation network for 3d
  shape recognition}. In \bibinfo{booktitle}{\emph{Proceedings of the IEEE/CVF
  International Conference on Computer Vision}}. \bibinfo{pages}{1--11}.
\newblock


\bibitem[Hegde et~al\mbox{.}(2023)]%
        {hegde2023clip}
\bibfield{author}{\bibinfo{person}{Deepti Hegde}, \bibinfo{person}{Jeya
  Maria~Jose Valanarasu}, {and} \bibinfo{person}{Vishal~M Patel}.}
  \bibinfo{year}{2023}\natexlab{}.
\newblock \showarticletitle{CLIP goes 3D: Leveraging Prompt Tuning for Language
  Grounded 3D Recognition}.
\newblock \bibinfo{journal}{\emph{arXiv preprint arXiv:2303.11313}}
  (\bibinfo{year}{2023}).
\newblock


\bibitem[Hu et~al\mbox{.}(2023)]%
        {hu2023you}
\bibfield{author}{\bibinfo{person}{Jie Hu}, \bibinfo{person}{Linyan Huang},
  \bibinfo{person}{Tianhe Ren}, \bibinfo{person}{Shengchuan Zhang},
  \bibinfo{person}{Rongrong Ji}, {and} \bibinfo{person}{Liujuan Cao}.}
  \bibinfo{year}{2023}\natexlab{}.
\newblock \showarticletitle{You Only Segment Once: Towards Real-Time Panoptic
  Segmentation}. In \bibinfo{booktitle}{\emph{Proceedings of the IEEE/CVF
  Conference on Computer Vision and Pattern Recognition}}.
  \bibinfo{pages}{17819--17829}.
\newblock


\bibitem[Jin et~al\mbox{.}(2023)]%
        {jin2023refclip}
\bibfield{author}{\bibinfo{person}{Lei Jin}, \bibinfo{person}{Gen Luo},
  \bibinfo{person}{Yiyi Zhou}, \bibinfo{person}{Xiaoshuai Sun},
  \bibinfo{person}{Guannan Jiang}, \bibinfo{person}{Annan Shu}, {and}
  \bibinfo{person}{Rongrong Ji}.} \bibinfo{year}{2023}\natexlab{}.
\newblock \showarticletitle{RefCLIP: A Universal Teacher for Weakly Supervised
  Referring Expression Comprehension}. In \bibinfo{booktitle}{\emph{Proceedings
  of the IEEE/CVF Conference on Computer Vision and Pattern Recognition}}.
  \bibinfo{pages}{2681--2690}.
\newblock


\bibitem[Ju et~al\mbox{.}(2022)]%
        {ju2022prompting}
\bibfield{author}{\bibinfo{person}{Chen Ju}, \bibinfo{person}{Tengda Han},
  \bibinfo{person}{Kunhao Zheng}, \bibinfo{person}{Ya Zhang}, {and}
  \bibinfo{person}{Weidi Xie}.} \bibinfo{year}{2022}\natexlab{}.
\newblock \showarticletitle{Prompting visual-language models for efficient
  video understanding}. In \bibinfo{booktitle}{\emph{Computer Vision--ECCV
  2022: 17th European Conference, Tel Aviv, Israel, October 23--27, 2022,
  Proceedings, Part XXXV}}. Springer, \bibinfo{pages}{105--124}.
\newblock


\bibitem[Lee et~al\mbox{.}(2022)]%
        {lee2022sagemix}
\bibfield{author}{\bibinfo{person}{Sanghyeok Lee}, \bibinfo{person}{Minkyu
  Jeon}, \bibinfo{person}{Injae Kim}, \bibinfo{person}{Yunyang Xiong}, {and}
  \bibinfo{person}{Hyunwoo~J Kim}.} \bibinfo{year}{2022}\natexlab{}.
\newblock \showarticletitle{Sagemix: Saliency-guided mixup for point clouds}.
\newblock \bibinfo{journal}{\emph{Advances in Neural Information Processing
  Systems}}  \bibinfo{volume}{35} (\bibinfo{year}{2022}),
  \bibinfo{pages}{23580--23592}.
\newblock


\bibitem[Li et~al\mbox{.}(2022a)]%
        {li2022align}
\bibfield{author}{\bibinfo{person}{Dongxu Li}, \bibinfo{person}{Junnan Li},
  \bibinfo{person}{Hongdong Li}, \bibinfo{person}{Juan~Carlos Niebles}, {and}
  \bibinfo{person}{Steven~CH Hoi}.} \bibinfo{year}{2022}\natexlab{a}.
\newblock \showarticletitle{Align and prompt: Video-and-language pre-training
  with entity prompts}. In \bibinfo{booktitle}{\emph{Proceedings of the
  IEEE/CVF Conference on Computer Vision and Pattern Recognition}}.
  \bibinfo{pages}{4953--4963}.
\newblock


\bibitem[Li et~al\mbox{.}(2019)]%
        {li2019visualbert}
\bibfield{author}{\bibinfo{person}{Liunian~Harold Li}, \bibinfo{person}{Mark
  Yatskar}, \bibinfo{person}{Da Yin}, \bibinfo{person}{Cho-Jui Hsieh}, {and}
  \bibinfo{person}{Kai-Wei Chang}.} \bibinfo{year}{2019}\natexlab{}.
\newblock \showarticletitle{Visualbert: A simple and performant baseline for
  vision and language}.
\newblock \bibinfo{journal}{\emph{arXiv preprint arXiv:1908.03557}}
  (\bibinfo{year}{2019}).
\newblock


\bibitem[Li et~al\mbox{.}(2022c)]%
        {li2022grounded}
\bibfield{author}{\bibinfo{person}{Liunian~Harold Li},
  \bibinfo{person}{Pengchuan Zhang}, \bibinfo{person}{Haotian Zhang},
  \bibinfo{person}{Jianwei Yang}, \bibinfo{person}{Chunyuan Li},
  \bibinfo{person}{Yiwu Zhong}, \bibinfo{person}{Lijuan Wang},
  \bibinfo{person}{Lu Yuan}, \bibinfo{person}{Lei Zhang},
  \bibinfo{person}{Jenq-Neng Hwang}, {et~al\mbox{.}}}
  \bibinfo{year}{2022}\natexlab{c}.
\newblock \showarticletitle{Grounded language-image pre-training}. In
  \bibinfo{booktitle}{\emph{Proceedings of the IEEE/CVF Conference on Computer
  Vision and Pattern Recognition}}. \bibinfo{pages}{10965--10975}.
\newblock


\bibitem[Li et~al\mbox{.}(2020)]%
        {li2020oscar}
\bibfield{author}{\bibinfo{person}{Xiujun Li}, \bibinfo{person}{Xi Yin},
  \bibinfo{person}{Chunyuan Li}, \bibinfo{person}{Pengchuan Zhang},
  \bibinfo{person}{Xiaowei Hu}, \bibinfo{person}{Lei Zhang},
  \bibinfo{person}{Lijuan Wang}, \bibinfo{person}{Houdong Hu},
  \bibinfo{person}{Li Dong}, \bibinfo{person}{Furu Wei}, {et~al\mbox{.}}}
  \bibinfo{year}{2020}\natexlab{}.
\newblock \showarticletitle{Oscar: Object-semantics aligned pre-training for
  vision-language tasks}. In \bibinfo{booktitle}{\emph{Computer Vision--ECCV
  2020: 16th European Conference, Glasgow, UK, August 23--28, 2020,
  Proceedings, Part XXX 16}}. Springer, \bibinfo{pages}{121--137}.
\newblock


\bibitem[Li et~al\mbox{.}(2022b)]%
        {li2022deepfusion}
\bibfield{author}{\bibinfo{person}{Yingwei Li}, \bibinfo{person}{Adams~Wei Yu},
  \bibinfo{person}{Tianjian Meng}, \bibinfo{person}{Ben Caine},
  \bibinfo{person}{Jiquan Ngiam}, \bibinfo{person}{Daiyi Peng},
  \bibinfo{person}{Junyang Shen}, \bibinfo{person}{Yifeng Lu},
  \bibinfo{person}{Denny Zhou}, \bibinfo{person}{Quoc~V Le}, {et~al\mbox{.}}}
  \bibinfo{year}{2022}\natexlab{b}.
\newblock \showarticletitle{Deepfusion: Lidar-camera deep fusion for
  multi-modal 3d object detection}. In \bibinfo{booktitle}{\emph{Proceedings of
  the IEEE/CVF Conference on Computer Vision and Pattern Recognition}}.
  \bibinfo{pages}{17182--17191}.
\newblock


\bibitem[Liu et~al\mbox{.}(2022)]%
        {liu2022masked}
\bibfield{author}{\bibinfo{person}{Haotian Liu}, \bibinfo{person}{Mu Cai},
  {and} \bibinfo{person}{Yong~Jae Lee}.} \bibinfo{year}{2022}\natexlab{}.
\newblock \showarticletitle{Masked discrimination for self-supervised learning
  on point clouds}. In \bibinfo{booktitle}{\emph{Computer Vision--ECCV 2022:
  17th European Conference, Tel Aviv, Israel, October 23--27, 2022,
  Proceedings, Part II}}. Springer, \bibinfo{pages}{657--675}.
\newblock


\bibitem[Liu et~al\mbox{.}(2019)]%
        {liu2019densepoint}
\bibfield{author}{\bibinfo{person}{Yongcheng Liu}, \bibinfo{person}{Bin Fan},
  \bibinfo{person}{Gaofeng Meng}, \bibinfo{person}{Jiwen Lu},
  \bibinfo{person}{Shiming Xiang}, {and} \bibinfo{person}{Chunhong Pan}.}
  \bibinfo{year}{2019}\natexlab{}.
\newblock \showarticletitle{Densepoint: Learning densely contextual
  representation for efficient point cloud processing}. In
  \bibinfo{booktitle}{\emph{Proceedings of the IEEE/CVF international
  conference on computer vision}}. \bibinfo{pages}{5239--5248}.
\newblock


\bibitem[Liu et~al\mbox{.}(2020)]%
        {liu2020closer}
\bibfield{author}{\bibinfo{person}{Ze Liu}, \bibinfo{person}{Han Hu},
  \bibinfo{person}{Yue Cao}, \bibinfo{person}{Zheng Zhang}, {and}
  \bibinfo{person}{Xin Tong}.} \bibinfo{year}{2020}\natexlab{}.
\newblock \showarticletitle{A closer look at local aggregation operators in
  point cloud analysis}. In \bibinfo{booktitle}{\emph{Computer Vision--ECCV
  2020: 16th European Conference, Glasgow, UK, August 23--28, 2020,
  Proceedings, Part XXIII 16}}. Springer, \bibinfo{pages}{326--342}.
\newblock


\bibitem[Liu et~al\mbox{.}(2021)]%
        {liu2021group}
\bibfield{author}{\bibinfo{person}{Ze Liu}, \bibinfo{person}{Zheng Zhang},
  \bibinfo{person}{Yue Cao}, \bibinfo{person}{Han Hu}, {and}
  \bibinfo{person}{Xin Tong}.} \bibinfo{year}{2021}\natexlab{}.
\newblock \showarticletitle{Group-free 3d object detection via transformers}.
  In \bibinfo{booktitle}{\emph{Proceedings of the IEEE/CVF International
  Conference on Computer Vision}}. \bibinfo{pages}{2949--2958}.
\newblock


\bibitem[Lu et~al\mbox{.}(2019)]%
        {lu2019vilbert}
\bibfield{author}{\bibinfo{person}{Jiasen Lu}, \bibinfo{person}{Dhruv Batra},
  \bibinfo{person}{Devi Parikh}, {and} \bibinfo{person}{Stefan Lee}.}
  \bibinfo{year}{2019}\natexlab{}.
\newblock \showarticletitle{Vilbert: Pretraining task-agnostic visiolinguistic
  representations for vision-and-language tasks}.
\newblock \bibinfo{journal}{\emph{Advances in neural information processing
  systems}}  \bibinfo{volume}{32} (\bibinfo{year}{2019}).
\newblock


\bibitem[Luo et~al\mbox{.}(2022)]%
        {luo2022clip4clip}
\bibfield{author}{\bibinfo{person}{Huaishao Luo}, \bibinfo{person}{Lei Ji},
  \bibinfo{person}{Ming Zhong}, \bibinfo{person}{Yang Chen},
  \bibinfo{person}{Wen Lei}, \bibinfo{person}{Nan Duan}, {and}
  \bibinfo{person}{Tianrui Li}.} \bibinfo{year}{2022}\natexlab{}.
\newblock \showarticletitle{CLIP4Clip: An empirical study of CLIP for end to
  end video clip retrieval and captioning}.
\newblock \bibinfo{journal}{\emph{Neurocomputing}}  \bibinfo{volume}{508}
  (\bibinfo{year}{2022}), \bibinfo{pages}{293--304}.
\newblock


\bibitem[Ma et~al\mbox{.}({[n.\,d.]})]%
        {marethinking}
\bibfield{author}{\bibinfo{person}{Xu Ma}, \bibinfo{person}{Can Qin},
  \bibinfo{person}{Haoxuan You}, \bibinfo{person}{Haoxi Ran}, {and}
  \bibinfo{person}{Yun Fu}.} \bibinfo{year}{[n.\,d.]}\natexlab{}.
\newblock \showarticletitle{Rethinking Network Design and Local Geometry in
  Point Cloud: A Simple Residual MLP Framework}. In
  \bibinfo{booktitle}{\emph{International Conference on Learning
  Representations}}.
\newblock


\bibitem[Ma et~al\mbox{.}(2022)]%
        {ma2022xclip}
\bibfield{author}{\bibinfo{person}{Yiwei Ma}, \bibinfo{person}{Guohai Xu},
  \bibinfo{person}{Xiaoshuai Sun}, \bibinfo{person}{Ming Yan},
  \bibinfo{person}{Ji Zhang}, {and} \bibinfo{person}{Rongrong Ji}.}
  \bibinfo{year}{2022}\natexlab{}.
\newblock \showarticletitle{X-clip: End-to-end multi-grained contrastive
  learning for video-text retrieval}. In \bibinfo{booktitle}{\emph{Proceedings
  of the 30th ACM International Conference on Multimedia}}.
  \bibinfo{pages}{638--647}.
\newblock


\bibitem[Ma et~al\mbox{.}(2023)]%
        {ma2023xmesh}
\bibfield{author}{\bibinfo{person}{Yiwei Ma}, \bibinfo{person}{Xiaioqing
  Zhang}, \bibinfo{person}{Xiaoshuai Sun}, \bibinfo{person}{Jiayi Ji},
  \bibinfo{person}{Haowei Wang}, \bibinfo{person}{Guannan Jiang},
  \bibinfo{person}{Weilin Zhuang}, {and} \bibinfo{person}{Rongrong Ji}.}
  \bibinfo{year}{2023}\natexlab{}.
\newblock \showarticletitle{X-Mesh: Towards Fast and Accurate Text-driven 3D
  Stylization via Dynamic Textual Guidance}.
\newblock \bibinfo{journal}{\emph{arXiv preprint arXiv:2303.15764}}
  (\bibinfo{year}{2023}).
\newblock


\bibitem[Maturana and Scherer(2015)]%
        {maturana2015voxnet}
\bibfield{author}{\bibinfo{person}{Daniel Maturana} {and}
  \bibinfo{person}{Sebastian Scherer}.} \bibinfo{year}{2015}\natexlab{}.
\newblock \showarticletitle{Voxnet: A 3d convolutional neural network for
  real-time object recognition}. In \bibinfo{booktitle}{\emph{2015 IEEE/RSJ
  international conference on intelligent robots and systems (IROS)}}. IEEE,
  \bibinfo{pages}{922--928}.
\newblock


\bibitem[Mu et~al\mbox{.}(2022)]%
        {mu2022slip}
\bibfield{author}{\bibinfo{person}{Norman Mu}, \bibinfo{person}{Alexander
  Kirillov}, \bibinfo{person}{David Wagner}, {and} \bibinfo{person}{Saining
  Xie}.} \bibinfo{year}{2022}\natexlab{}.
\newblock \showarticletitle{Slip: Self-supervision meets language-image
  pre-training}. In \bibinfo{booktitle}{\emph{Computer Vision--ECCV 2022: 17th
  European Conference, Tel Aviv, Israel, October 23--27, 2022, Proceedings,
  Part XXVI}}. Springer, \bibinfo{pages}{529--544}.
\newblock


\bibitem[Pang et~al\mbox{.}(2022)]%
        {pang2022masked}
\bibfield{author}{\bibinfo{person}{Yatian Pang}, \bibinfo{person}{Wenxiao
  Wang}, \bibinfo{person}{Francis~EH Tay}, \bibinfo{person}{Wei Liu},
  \bibinfo{person}{Yonghong Tian}, {and} \bibinfo{person}{Li Yuan}.}
  \bibinfo{year}{2022}\natexlab{}.
\newblock \showarticletitle{Masked autoencoders for point cloud self-supervised
  learning}. In \bibinfo{booktitle}{\emph{Computer Vision--ECCV 2022: 17th
  European Conference, Tel Aviv, Israel, October 23--27, 2022, Proceedings,
  Part II}}. Springer, \bibinfo{pages}{604--621}.
\newblock


\bibitem[Qi et~al\mbox{.}(2017a)]%
        {qi2017pointnet}
\bibfield{author}{\bibinfo{person}{Charles~R Qi}, \bibinfo{person}{Hao Su},
  \bibinfo{person}{Kaichun Mo}, {and} \bibinfo{person}{Leonidas~J Guibas}.}
  \bibinfo{year}{2017}\natexlab{a}.
\newblock \showarticletitle{Pointnet: Deep learning on point sets for 3d
  classification and segmentation}. In \bibinfo{booktitle}{\emph{Proceedings of
  the IEEE conference on computer vision and pattern recognition}}.
  \bibinfo{pages}{652--660}.
\newblock


\bibitem[Qi et~al\mbox{.}(2017b)]%
        {qi2017pointnet++}
\bibfield{author}{\bibinfo{person}{Charles~Ruizhongtai Qi}, \bibinfo{person}{Li
  Yi}, \bibinfo{person}{Hao Su}, {and} \bibinfo{person}{Leonidas~J Guibas}.}
  \bibinfo{year}{2017}\natexlab{b}.
\newblock \showarticletitle{Pointnet++: Deep hierarchical feature learning on
  point sets in a metric space}.
\newblock \bibinfo{journal}{\emph{Advances in neural information processing
  systems}}  \bibinfo{volume}{30} (\bibinfo{year}{2017}).
\newblock


\bibitem[Qian et~al\mbox{.}(2022)]%
        {qian2022pointnext}
\bibfield{author}{\bibinfo{person}{Guocheng Qian}, \bibinfo{person}{Yuchen Li},
  \bibinfo{person}{Houwen Peng}, \bibinfo{person}{Jinjie Mai},
  \bibinfo{person}{Hasan Hammoud}, \bibinfo{person}{Mohamed Elhoseiny}, {and}
  \bibinfo{person}{Bernard Ghanem}.} \bibinfo{year}{2022}\natexlab{}.
\newblock \showarticletitle{Pointnext: Revisiting pointnet++ with improved
  training and scaling strategies}.
\newblock \bibinfo{journal}{\emph{Advances in Neural Information Processing
  Systems}}  \bibinfo{volume}{35} (\bibinfo{year}{2022}),
  \bibinfo{pages}{23192--23204}.
\newblock


\bibitem[Radford et~al\mbox{.}(2021)]%
        {radford2021learning}
\bibfield{author}{\bibinfo{person}{Alec Radford}, \bibinfo{person}{Jong~Wook
  Kim}, \bibinfo{person}{Chris Hallacy}, \bibinfo{person}{Aditya Ramesh},
  \bibinfo{person}{Gabriel Goh}, \bibinfo{person}{Sandhini Agarwal},
  \bibinfo{person}{Girish Sastry}, \bibinfo{person}{Amanda Askell},
  \bibinfo{person}{Pamela Mishkin}, \bibinfo{person}{Jack Clark},
  {et~al\mbox{.}}} \bibinfo{year}{2021}\natexlab{}.
\newblock \showarticletitle{Learning transferable visual models from natural
  language supervision}. In \bibinfo{booktitle}{\emph{International conference
  on machine learning}}. PMLR, \bibinfo{pages}{8748--8763}.
\newblock


\bibitem[Ran et~al\mbox{.}(2022)]%
        {ran2022surface}
\bibfield{author}{\bibinfo{person}{Haoxi Ran}, \bibinfo{person}{Jun Liu}, {and}
  \bibinfo{person}{Chengjie Wang}.} \bibinfo{year}{2022}\natexlab{}.
\newblock \showarticletitle{Surface representation for point clouds}. In
  \bibinfo{booktitle}{\emph{Proceedings of the IEEE/CVF Conference on Computer
  Vision and Pattern Recognition}}. \bibinfo{pages}{18942--18952}.
\newblock


\bibitem[Shi et~al\mbox{.}(2020)]%
        {shi2020pv}
\bibfield{author}{\bibinfo{person}{Shaoshuai Shi}, \bibinfo{person}{Chaoxu
  Guo}, \bibinfo{person}{Li Jiang}, \bibinfo{person}{Zhe Wang},
  \bibinfo{person}{Jianping Shi}, \bibinfo{person}{Xiaogang Wang}, {and}
  \bibinfo{person}{Hongsheng Li}.} \bibinfo{year}{2020}\natexlab{}.
\newblock \showarticletitle{Pv-rcnn: Point-voxel feature set abstraction for 3d
  object detection}. In \bibinfo{booktitle}{\emph{Proceedings of the IEEE/CVF
  Conference on Computer Vision and Pattern Recognition}}.
  \bibinfo{pages}{10529--10538}.
\newblock


\bibitem[Sun et~al\mbox{.}(2009)]%
        {sun2009concise}
\bibfield{author}{\bibinfo{person}{Jian Sun}, \bibinfo{person}{Maks
  Ovsjanikov}, {and} \bibinfo{person}{Leonidas Guibas}.}
  \bibinfo{year}{2009}\natexlab{}.
\newblock \showarticletitle{A concise and provably informative multi-scale
  signature based on heat diffusion}. In \bibinfo{booktitle}{\emph{Computer
  graphics forum}}, Vol.~\bibinfo{volume}{28}. Wiley Online Library,
  \bibinfo{pages}{1383--1392}.
\newblock


\bibitem[Tan and Bansal(2019)]%
        {tan2019lxmert}
\bibfield{author}{\bibinfo{person}{Hao Tan} {and} \bibinfo{person}{Mohit
  Bansal}.} \bibinfo{year}{2019}\natexlab{}.
\newblock \showarticletitle{LXMERT: Learning Cross-Modality Encoder
  Representations from Transformers}. In \bibinfo{booktitle}{\emph{Proceedings
  of the 2019 Conference on Empirical Methods in Natural Language Processing
  and the 9th International Joint Conference on Natural Language Processing
  (EMNLP-IJCNLP)}}. \bibinfo{pages}{5100--5111}.
\newblock


\bibitem[Tolstikhin et~al\mbox{.}(2021)]%
        {tolstikhin2021mlp}
\bibfield{author}{\bibinfo{person}{Ilya~O Tolstikhin}, \bibinfo{person}{Neil
  Houlsby}, \bibinfo{person}{Alexander Kolesnikov}, \bibinfo{person}{Lucas
  Beyer}, \bibinfo{person}{Xiaohua Zhai}, \bibinfo{person}{Thomas Unterthiner},
  \bibinfo{person}{Jessica Yung}, \bibinfo{person}{Andreas Steiner},
  \bibinfo{person}{Daniel Keysers}, \bibinfo{person}{Jakob Uszkoreit},
  {et~al\mbox{.}}} \bibinfo{year}{2021}\natexlab{}.
\newblock \showarticletitle{Mlp-mixer: An all-mlp architecture for vision}.
\newblock \bibinfo{journal}{\emph{Advances in neural information processing
  systems}}  \bibinfo{volume}{34} (\bibinfo{year}{2021}),
  \bibinfo{pages}{24261--24272}.
\newblock


\bibitem[Uy et~al\mbox{.}(2019)]%
        {uy2019revisiting}
\bibfield{author}{\bibinfo{person}{Mikaela~Angelina Uy},
  \bibinfo{person}{Quang-Hieu Pham}, \bibinfo{person}{Binh-Son Hua},
  \bibinfo{person}{Thanh Nguyen}, {and} \bibinfo{person}{Sai-Kit Yeung}.}
  \bibinfo{year}{2019}\natexlab{}.
\newblock \showarticletitle{Revisiting point cloud classification: A new
  benchmark dataset and classification model on real-world data}. In
  \bibinfo{booktitle}{\emph{Proceedings of the IEEE/CVF international
  conference on computer vision}}. \bibinfo{pages}{1588--1597}.
\newblock


\bibitem[Vaswani et~al\mbox{.}(2017)]%
        {vaswani2017attention}
\bibfield{author}{\bibinfo{person}{Ashish Vaswani}, \bibinfo{person}{Noam
  Shazeer}, \bibinfo{person}{Niki Parmar}, \bibinfo{person}{Jakob Uszkoreit},
  \bibinfo{person}{Llion Jones}, \bibinfo{person}{Aidan~N Gomez},
  \bibinfo{person}{{\L}ukasz Kaiser}, {and} \bibinfo{person}{Illia
  Polosukhin}.} \bibinfo{year}{2017}\natexlab{}.
\newblock \showarticletitle{Attention is all you need}.
\newblock \bibinfo{journal}{\emph{Advances in neural information processing
  systems}}  \bibinfo{volume}{30} (\bibinfo{year}{2017}).
\newblock


\bibitem[Vu et~al\mbox{.}(2022)]%
        {vu2022softgroup}
\bibfield{author}{\bibinfo{person}{Thang Vu}, \bibinfo{person}{Kookhoi Kim},
  \bibinfo{person}{Tung~M Luu}, \bibinfo{person}{Thanh Nguyen}, {and}
  \bibinfo{person}{Chang~D Yoo}.} \bibinfo{year}{2022}\natexlab{}.
\newblock \showarticletitle{Softgroup for 3d instance segmentation on point
  clouds}. In \bibinfo{booktitle}{\emph{Proceedings of the IEEE/CVF Conference
  on Computer Vision and Pattern Recognition}}. \bibinfo{pages}{2708--2717}.
\newblock


\bibitem[Wang et~al\mbox{.}(2023)]%
        {wang2023towards}
\bibfield{author}{\bibinfo{person}{Haowei Wang}, \bibinfo{person}{Jiayi Ji},
  \bibinfo{person}{Yiyi Zhou}, \bibinfo{person}{Yongjian Wu}, {and}
  \bibinfo{person}{Xiaoshuai Sun}.} \bibinfo{year}{2023}\natexlab{}.
\newblock \showarticletitle{Towards real-time panoptic narrative grounding by
  an end-to-end grounding network}.
\newblock \bibinfo{journal}{\emph{arXiv preprint arXiv:2301.03160}}
  (\bibinfo{year}{2023}).
\newblock


\bibitem[Wang et~al\mbox{.}(2022)]%
        {wang2022p2p}
\bibfield{author}{\bibinfo{person}{Ziyi Wang}, \bibinfo{person}{Xumin Yu},
  \bibinfo{person}{Yongming Rao}, \bibinfo{person}{Jie Zhou}, {and}
  \bibinfo{person}{Jiwen Lu}.} \bibinfo{year}{2022}\natexlab{}.
\newblock \showarticletitle{P2p: Tuning pre-trained image models for point
  cloud analysis with point-to-pixel prompting}.
\newblock \bibinfo{journal}{\emph{Advances in neural information processing
  systems}}  \bibinfo{volume}{35} (\bibinfo{year}{2022}),
  \bibinfo{pages}{14388--14402}.
\newblock


\bibitem[Wu et~al\mbox{.}(2018)]%
        {wu2018dgcnn}
\bibfield{author}{\bibinfo{person}{Bo Wu}, \bibinfo{person}{Yang Liu},
  \bibinfo{person}{Bo Lang}, {and} \bibinfo{person}{Lei Huang}.}
  \bibinfo{year}{2018}\natexlab{}.
\newblock \showarticletitle{Dgcnn: Disordered graph convolutional neural
  network based on the gaussian mixture model}.
\newblock \bibinfo{journal}{\emph{Neurocomputing}}  \bibinfo{volume}{321}
  (\bibinfo{year}{2018}), \bibinfo{pages}{346--356}.
\newblock


\bibitem[Wu et~al\mbox{.}(2015)]%
        {wu20153d}
\bibfield{author}{\bibinfo{person}{Zhirong Wu}, \bibinfo{person}{Shuran Song},
  \bibinfo{person}{Aditya Khosla}, \bibinfo{person}{Fisher Yu},
  \bibinfo{person}{Linguang Zhang}, \bibinfo{person}{Xiaoou Tang}, {and}
  \bibinfo{person}{Jianxiong Xiao}.} \bibinfo{year}{2015}\natexlab{}.
\newblock \showarticletitle{3d shapenets: A deep representation for volumetric
  shapes}. In \bibinfo{booktitle}{\emph{Proceedings of the IEEE conference on
  computer vision and pattern recognition}}. \bibinfo{pages}{1912--1920}.
\newblock


\bibitem[Xiao et~al\mbox{.}(2023)]%
        {xiao2023unsupervised}
\bibfield{author}{\bibinfo{person}{Aoran Xiao}, \bibinfo{person}{Jiaxing
  Huang}, \bibinfo{person}{Dayan Guan}, \bibinfo{person}{Xiaoqin Zhang},
  \bibinfo{person}{Shijian Lu}, {and} \bibinfo{person}{Ling Shao}.}
  \bibinfo{year}{2023}\natexlab{}.
\newblock \showarticletitle{Unsupervised Point Cloud Representation Learning
  with Deep Neural Networks: A Survey}.
\newblock \bibinfo{journal}{\emph{IEEE Transactions on Pattern Analysis and
  Machine Intelligence}} (\bibinfo{year}{2023}).
\newblock


\bibitem[Xie et~al\mbox{.}(2020)]%
        {xie2020grnet}
\bibfield{author}{\bibinfo{person}{Haozhe Xie}, \bibinfo{person}{Hongxun Yao},
  \bibinfo{person}{Shangchen Zhou}, \bibinfo{person}{Jiageng Mao},
  \bibinfo{person}{Shengping Zhang}, {and} \bibinfo{person}{Wenxiu Sun}.}
  \bibinfo{year}{2020}\natexlab{}.
\newblock \showarticletitle{Grnet: Gridding residual network for dense point
  cloud completion}. In \bibinfo{booktitle}{\emph{Computer Vision--ECCV 2020:
  16th European Conference, Glasgow, UK, August 23--28, 2020, Proceedings, Part
  IX}}. Springer, \bibinfo{pages}{365--381}.
\newblock


\bibitem[Xu et~al\mbox{.}(2021b)]%
        {xu2021videoclip}
\bibfield{author}{\bibinfo{person}{Hu Xu}, \bibinfo{person}{Gargi Ghosh},
  \bibinfo{person}{Po-Yao Huang}, \bibinfo{person}{Dmytro Okhonko},
  \bibinfo{person}{Armen Aghajanyan}, \bibinfo{person}{Florian Metze},
  \bibinfo{person}{Luke Zettlemoyer}, {and} \bibinfo{person}{Christoph
  Feichtenhofer}.} \bibinfo{year}{2021}\natexlab{b}.
\newblock \showarticletitle{VideoCLIP: Contrastive Pre-training for Zero-shot
  Video-Text Understanding}. In \bibinfo{booktitle}{\emph{Proceedings of the
  2021 Conference on Empirical Methods in Natural Language Processing}}.
  \bibinfo{pages}{6787--6800}.
\newblock


\bibitem[Xu et~al\mbox{.}(2021a)]%
        {xu2021paconv}
\bibfield{author}{\bibinfo{person}{Mutian Xu}, \bibinfo{person}{Runyu Ding},
  \bibinfo{person}{Hengshuang Zhao}, {and} \bibinfo{person}{Xiaojuan Qi}.}
  \bibinfo{year}{2021}\natexlab{a}.
\newblock \showarticletitle{Paconv: Position adaptive convolution with dynamic
  kernel assembling on point clouds}. In \bibinfo{booktitle}{\emph{Proceedings
  of the IEEE/CVF Conference on Computer Vision and Pattern Recognition}}.
  \bibinfo{pages}{3173--3182}.
\newblock


\bibitem[Xue et~al\mbox{.}(2022)]%
        {xue2022ulip}
\bibfield{author}{\bibinfo{person}{Le Xue}, \bibinfo{person}{Mingfei Gao},
  \bibinfo{person}{Chen Xing}, \bibinfo{person}{Roberto
  Mart{\'\i}n-Mart{\'\i}n}, \bibinfo{person}{Jiajun Wu},
  \bibinfo{person}{Caiming Xiong}, \bibinfo{person}{Ran Xu},
  \bibinfo{person}{Juan~Carlos Niebles}, {and} \bibinfo{person}{Silvio
  Savarese}.} \bibinfo{year}{2022}\natexlab{}.
\newblock \showarticletitle{ULIP: Learning Unified Representation of Language,
  Image and Point Cloud for 3D Understanding}.
\newblock \bibinfo{journal}{\emph{arXiv preprint arXiv:2212.05171}}
  (\bibinfo{year}{2022}).
\newblock


\bibitem[Yan et~al\mbox{.}({[n.\,d.]})]%
        {yanlet}
\bibfield{author}{\bibinfo{person}{Xu Yan}, \bibinfo{person}{Heshen Zhan},
  \bibinfo{person}{Chaoda Zheng}, \bibinfo{person}{Jiantao Gao},
  \bibinfo{person}{Ruimao Zhang}, \bibinfo{person}{Shuguang Cui}, {and}
  \bibinfo{person}{Zhen Li}.} \bibinfo{year}{[n.\,d.]}\natexlab{}.
\newblock \showarticletitle{Let Images Give You More: Point Cloud Cross-Modal
  Training for Shape Analysis}. In \bibinfo{booktitle}{\emph{Advances in Neural
  Information Processing Systems}}.
\newblock


\bibitem[Yin et~al\mbox{.}(2021)]%
        {yin2021center}
\bibfield{author}{\bibinfo{person}{Tianwei Yin}, \bibinfo{person}{Xingyi Zhou},
  {and} \bibinfo{person}{Philipp Krahenbuhl}.} \bibinfo{year}{2021}\natexlab{}.
\newblock \showarticletitle{Center-based 3d object detection and tracking}. In
  \bibinfo{booktitle}{\emph{Proceedings of the IEEE/CVF conference on computer
  vision and pattern recognition}}. \bibinfo{pages}{11784--11793}.
\newblock


\bibitem[Yu et~al\mbox{.}(2022)]%
        {yu2022point}
\bibfield{author}{\bibinfo{person}{Xumin Yu}, \bibinfo{person}{Lulu Tang},
  \bibinfo{person}{Yongming Rao}, \bibinfo{person}{Tiejun Huang},
  \bibinfo{person}{Jie Zhou}, {and} \bibinfo{person}{Jiwen Lu}.}
  \bibinfo{year}{2022}\natexlab{}.
\newblock \showarticletitle{Point-bert: Pre-training 3d point cloud
  transformers with masked point modeling}. In
  \bibinfo{booktitle}{\emph{Proceedings of the IEEE/CVF Conference on Computer
  Vision and Pattern Recognition}}. \bibinfo{pages}{19313--19322}.
\newblock


\bibitem[Zhang et~al\mbox{.}(2022b)]%
        {zhang2022glipv2}
\bibfield{author}{\bibinfo{person}{Haotian Zhang}, \bibinfo{person}{Pengchuan
  Zhang}, \bibinfo{person}{Xiaowei Hu}, \bibinfo{person}{Yen-Chun Chen},
  \bibinfo{person}{Liunian Li}, \bibinfo{person}{Xiyang Dai},
  \bibinfo{person}{Lijuan Wang}, \bibinfo{person}{Lu Yuan},
  \bibinfo{person}{Jenq-Neng Hwang}, {and} \bibinfo{person}{Jianfeng Gao}.}
  \bibinfo{year}{2022}\natexlab{b}.
\newblock \showarticletitle{Glipv2: Unifying localization and vision-language
  understanding}.
\newblock \bibinfo{journal}{\emph{Advances in Neural Information Processing
  Systems}}  \bibinfo{volume}{35} (\bibinfo{year}{2022}),
  \bibinfo{pages}{36067--36080}.
\newblock


\bibitem[Zhang et~al\mbox{.}(2023)]%
        {zhang2023clip}
\bibfield{author}{\bibinfo{person}{Junbo Zhang}, \bibinfo{person}{Runpei Dong},
  {and} \bibinfo{person}{Kaisheng Ma}.} \bibinfo{year}{2023}\natexlab{}.
\newblock \showarticletitle{CLIP-FO3D: Learning Free Open-world 3D Scene
  Representations from 2D Dense CLIP}.
\newblock \bibinfo{journal}{\emph{arXiv preprint arXiv:2303.04748}}
  (\bibinfo{year}{2023}).
\newblock


\bibitem[Zhang et~al\mbox{.}({[n.\,d.]})]%
        {zhangpoint}
\bibfield{author}{\bibinfo{person}{Renrui Zhang}, \bibinfo{person}{Ziyu Guo},
  \bibinfo{person}{Peng Gao}, \bibinfo{person}{Rongyao Fang},
  \bibinfo{person}{Bin Zhao}, \bibinfo{person}{Dong Wang}, \bibinfo{person}{Yu
  Qiao}, {and} \bibinfo{person}{Hongsheng Li}.}
  \bibinfo{year}{[n.\,d.]}\natexlab{}.
\newblock \showarticletitle{Point-M2AE: Multi-scale Masked Autoencoders for
  Hierarchical Point Cloud Pre-training}. In \bibinfo{booktitle}{\emph{Advances
  in Neural Information Processing Systems}}.
\newblock


\bibitem[Zhang et~al\mbox{.}(2022a)]%
        {zhang2022pointclip}
\bibfield{author}{\bibinfo{person}{Renrui Zhang}, \bibinfo{person}{Ziyu Guo},
  \bibinfo{person}{Wei Zhang}, \bibinfo{person}{Kunchang Li},
  \bibinfo{person}{Xupeng Miao}, \bibinfo{person}{Bin Cui}, \bibinfo{person}{Yu
  Qiao}, \bibinfo{person}{Peng Gao}, {and} \bibinfo{person}{Hongsheng Li}.}
  \bibinfo{year}{2022}\natexlab{a}.
\newblock \showarticletitle{Pointclip: Point cloud understanding by clip}. In
  \bibinfo{booktitle}{\emph{Proceedings of the IEEE/CVF Conference on Computer
  Vision and Pattern Recognition}}. \bibinfo{pages}{8552--8562}.
\newblock


\bibitem[Zhao et~al\mbox{.}(2021)]%
        {zhao2021point}
\bibfield{author}{\bibinfo{person}{Hengshuang Zhao}, \bibinfo{person}{Li
  Jiang}, \bibinfo{person}{Jiaya Jia}, \bibinfo{person}{Philip~HS Torr}, {and}
  \bibinfo{person}{Vladlen Koltun}.} \bibinfo{year}{2021}\natexlab{}.
\newblock \showarticletitle{Point transformer}. In
  \bibinfo{booktitle}{\emph{Proceedings of the IEEE/CVF international
  conference on computer vision}}. \bibinfo{pages}{16259--16268}.
\newblock


\bibitem[Zhao et~al\mbox{.}(2023)]%
        {zhao-etal-2023-generating-visual}
\bibfield{author}{\bibinfo{person}{Yu Zhao}, \bibinfo{person}{Hao Fei},
  \bibinfo{person}{Wei Ji}, \bibinfo{person}{Jianguo Wei},
  \bibinfo{person}{Meishan Zhang}, \bibinfo{person}{Min Zhang}, {and}
  \bibinfo{person}{Tat-Seng Chua}.} \bibinfo{year}{2023}\natexlab{}.
\newblock \showarticletitle{Generating Visual Spatial Description via Holistic
  3{D} Scene Understanding}. In \bibinfo{booktitle}{\emph{Proceedings of the
  61st Annual Meeting of the Association for Computational Linguistics (Volume
  1: Long Papers)}}. \bibinfo{pages}{7960--7977}.
\newblock


\end{thebibliography}

\clearpage

\appendix

\begin{table}[htb]
    \small
    \centering
        \caption{The semantic segmentation on S3DIS.}
    \begin{tabular}{lcc}
    \toprule
         Model& Overall Acc & Class avg IoU \\
         \midrule
         PointNet++ (ssg) \cite{qi2017pointnet++} &  83.0 & 53.5 \\
         PointNet++ (ssg) \cite{qi2017pointnet++} + JM3D &  \textbf{83.6} & 
 \textbf{57.1}\\
         \bottomrule
    \end{tabular}

    \label{tab:seg-semantic}
\end{table}

\begin{table}[htb]
    \small
    \centering
    \caption{The Part segmentation on ShapeNet.}
    \begin{tabular}{lcc}
    \toprule
         Model& Instance avg IoU & Class avg IoU \\
         \midrule
         SageMix \cite{lee2022sagemix} &  85.4 & - \\
         PointNet++ (ssg) \cite{qi2017pointnet++} &  84.9 & 81.8 \\
         PointNet++ (ssg) \cite{qi2017pointnet++} + JM3D &  \textbf{85.5} & 
 \textbf{82.1}\\
         \bottomrule
    \end{tabular}
    \label{tab:seg-part}
\end{table}

\begin{table}[htb]
    \caption{3D classification results on ScanObjectNN. We follow the default settings of the original method to train on the hardest set. JM3D achieves a significant improvement compared to the previous method, helping to improve the original backbone by 3.5\%.}
    \centering
    \begin{tabular}{lcc}
    \toprule
         Model& Overall Acc & Class-mean Acc \\
         \midrule
         PointNet \cite{qi2017pointnet} &  68.2 & 63.4 \\
         PointNet++ \cite{qi2017pointnet++} &  77.9 & 75.4 \\
         DGCNN \cite{wu2018dgcnn} &  78.1 & 73.6 \\
         MVTN \cite{hamdi2021mvtn} &  82.8 &  --\\
         PointBERT \cite{yu2022point} &  83.1 &  --\\
         RepSurf-U \cite{ran2022surface} & 84.6 &  --\\
         PointMAE \cite{pang2022masked} & 85.2 &  --\\
         RepSurf-U (2x) \cite{ran2022surface} &  86.0 &  --\\

         P2P \cite{wang2022p2p} & 89.3 & --\\
         \midrule
         PointMLP \cite{marethinking} &  85.7 & 84.4 \\
         PointMLP+ ULIP &  88.8 & 87.8 \\
         PointMLP + JM3D & 89.2 & 88.4 \\
         PointMLP$^{*}$ + JM3D & \textbf{89.5} & \textbf{88.7} \\
         \bottomrule
    \end{tabular}
    \label{tab:fintune-scan}
\end{table}

\section{3D classification Fine-tuning}
In order to show the potential of JM3D, we conduct finr-tuning experiments on ScanObjectNN based on the one of the SOTA frameworks, PointMLP~\cite{marethinking}.

During fine-tune stage, we only train the 3D encoder of JM3D on the \emph{PB\_T50\_RS} set of ScanObjectNN. This is because that the \emph{PB\_T50\_RS} set is a tough split of real-world scanned object with the background noise. We fine-tune PointMLP with the learing rate of 0.03, and weight decay is 3e-4 for 350 epochs.We directly hot start the pre-trained parameters and conduct fine-tuning entirely under the original model's conditions. We adhere to the standard practice in the research community by utilizing OA (Overall Accuracy) and mAcc (Class Average Accuracy) as our evaluation metrics. 

As shown on Tab.~\ref{tab:fintune-scan}, The performance of the backbone model is significantly improved with the assistance of JM3D. Specifically, JM3D takes 3.5\% improvement on PointMLP on OA and 4.0\% on mAcc. In addition, PointMLP + JM3D, outperforms the previous SOTA method RepSurf-U ($2\times$) by 3.2\%. With the voting stategy, PointMLP$^{*}$ + JM3D sets a new SOTA. Compared to the training method used in ULIP, JM3D outperformed it, indicating that JM3D has remarkable capabilities in directly enhancing existing models without any specially designed structures. 

\section{Details of Sets on ModelNet40 in zero-shot Classification}

As mentioned in the main text, we notice the shared categories between Shapenet55 and ModelNet40. To ensure a fairer comparison, we exclude similar categories and construct different validate sets for ModelNet40.

\noindent\textbf{All set}: all the categroyies in ModelNet40, which is shown in Tab.~\ref{tab:ModelNet40-All-Set}

\begin{table}[htb]
    \small
    \caption{ModelNet40 All Set.}
    \begin{tabular}{ccccc}
        \toprule
        airplane & bathtub & bed & bench & bookshelf \\
        \midrule
        bottle & bowl & car & chair & cone \\
        \midrule
        cup& curtain& desk& door& dresser \\
        \midrule
        flower\_pot& glass\_box& guitar& keyboard& lamp \\
        \midrule
        laptop& mantel& monitor& night\_stand& person \\
        \midrule
        piano& plant& radio& range\_hood& sink \\
        \midrule
        sofa& stairs& stool& table& tent \\
        \midrule
        toilet& tv\_stand& vase& wardrobe& xbox \\
        \bottomrule
    \end{tabular}
    \label{tab:ModelNet40-All-Set}
\end{table}

\noindent\textbf{Medium Set}: We exclud categories that have exactly the same names as those in our pre-training dataset ShapeNet55. The categories remaining in this set are displayed in Tab.~\ref{tab:ModelNet40-Medium-Set}.

\begin{table}[htb]
    \small
    \caption{ModelNet40 Medium Set.}
    \begin{tabular}{ccccc}
        \toprule
        cone& cup& curtain& door& dresser \\
        \midrule
        glass\_box& mantel& monitor& night\_stand& person \\
        \midrule
        plant& radio& range\_hood& sink& stairs \\
        \midrule
        stool& tent& toilet& tv\_stand& vase \\
        \midrule
        wardrobe& xbox \\
        \bottomrule
    \end{tabular}

    \label{tab:ModelNet40-Medium-Set}
\end{table}

\noindent\textbf{Hard Set}: Both the exact category names and their synonyms present in ShapeNet55 have been removed. The resulting set, known as the \emph{Hard Set}, is presented in Tab.~\ref{tab:ModelNet40-Hard-Set}.

\begin{table}[htb]
    \small
        \caption{ModelNet40 Hard Set.}
    \begin{tabular}{ccccc}
        \toprule
        cone& curtain& door& dresser& glass\_box \\
        \midrule
        mantel& night\_stand& person& plant& radio \\
        \midrule
        range\_hood& sink& stairs& tent& toilet \\
        \midrule
        tv\_stand& xbox \\
        \bottomrule
    \end{tabular}

    \label{tab:ModelNet40-Hard-Set}
\end{table}

\begin{figure*}[t]
\centering
\includegraphics[width=1.0\textwidth]{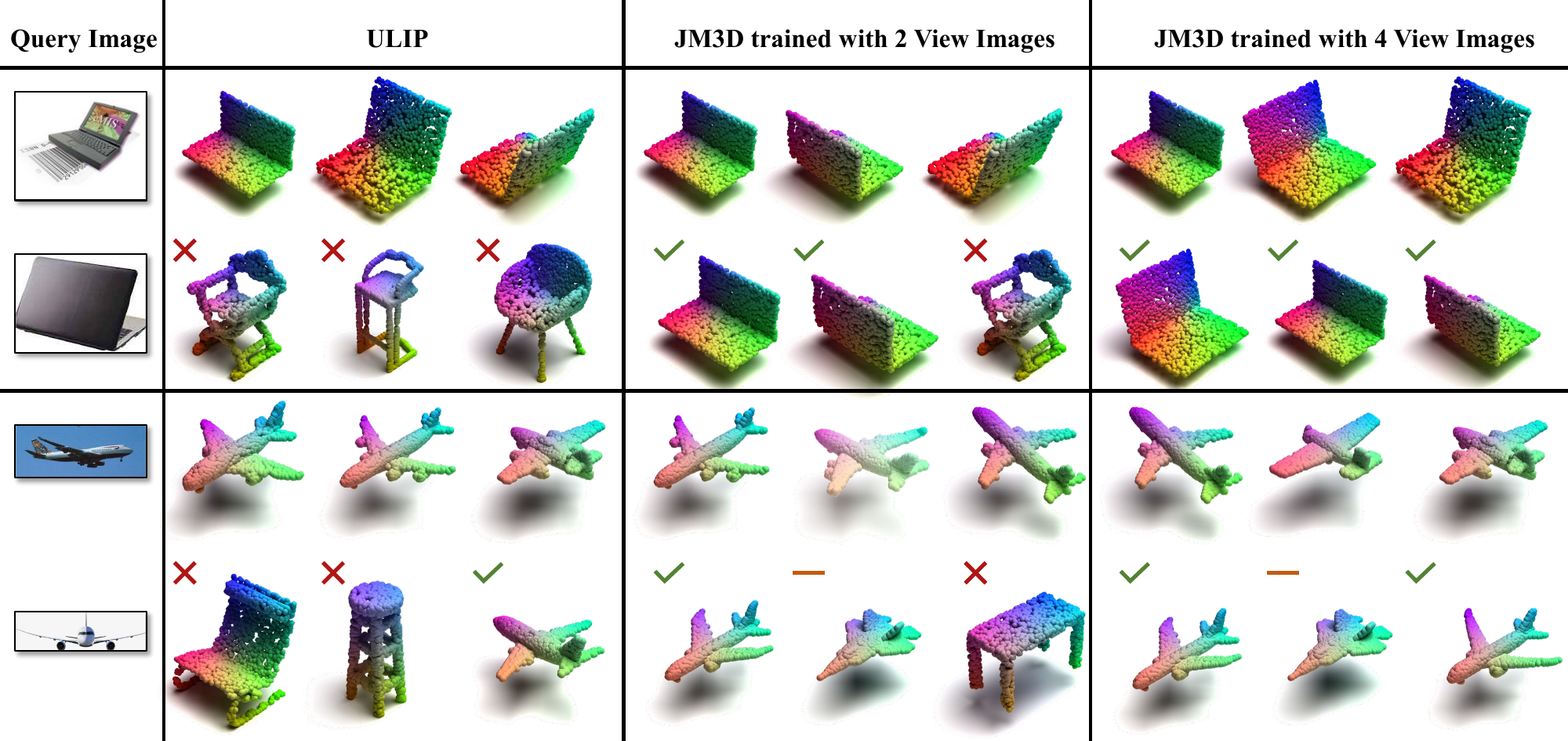}
\caption{The qualitative results of the real image to point cloud retrieval. Giving an image, We show the top-3 point cloud retrieval results from ModelNet40. All models perform well on the simple samples (the 1st row and the 3rd row). However, when it comes to the challenging samples (the 2nd row and the 4th row), JM3D demonstrates a more accurate retrieval ability compared to the previous state-of-the-art (ULIP). The JM3D trained with 4 view images shows better performance compared with the 2 view images, benefiting the more solid bias of vision modality.}
\label{fig3}
\end{figure*}

\section{Cross-modal Retrieval}\label{sec:cross-retrival}

Another noteworthy aspect is that JM3D confers enhanced cross-modal capabilities to the foundational point cloud model. In addition to the conducted quantitative analysis aligned with the language modality above, this section primarily showcases the qualitative results of image interaction ability enabled by JM3D."

We collect some images from a real-world dataset, \emph{i.e.}, Caltech101~\cite{fei2004learning} to retrieve 3D models from the ModelNet40 test set, which is a medium-scale dataset with more than 2.5K models across 40 categories. Meanwhile, some more challenging samples are constructed, which often possess unique perspectives, making them difficult to be recognized by conventional models. The top-3 retrieval models are presented in Fig.~\ref{fig3}.

In Fig.~\ref{fig3}, samples are belonging to the categories of "airplane" and "laptop", with each category further divided into two levels: a challenging level (top) and a simple level (bottom). It is apparent that for simplistic samples, all models exhibit a significant level of retrieving. Nonetheless, when the image is captured from an uncommon perspective, ULIP is unable to identify the appropriate point cloud. In contrast, JM3D trained with two views can identify some of the correct outcomes, whereas, with an increase in the number of images to 4 in the CIS, JM3D can successfully locate almost all of the appropriate models. The visualizations show an excellent sign that our model has learned meaningful features across visual and 3D modalities. Furthermore, this suggests that although an increase in the number of views in the CIS may have a minor impact on the text-based performance as Tab.~\ref{tab:ablation-image}, it can greatly enhance the alignment capability of the model in the image domain.

\section{Semantic Segmentation and Part Segmentation}

we have conducted semantic segmentation experiments on the S3DIS~\cite{armeni20163d} dataset, and partial segmentation experiments on the ShapeNet~\cite{chang2015shapenet}, the results of which are recorded in Tab.~\ref{tab:seg-semantic} and Tab.~\ref{tab:seg-part}. Our proposed method has continued to demonstrate great performance, achieving scores of 57.1 and 82.1 on the S3DIS and ShapeNet datasets respectively, and correspondingly increasing by 3.6 and 0.3 points. These scene understanding tasks indisputably demonstrate the effectiveness of our approach.

\end{document}